\documentclass{article}

\usepackage[final,nonatbib]{neurips_2024}

\usepackage[utf8]{inputenc} % allow utf-8 input
\usepackage[T1]{fontenc}    % use 8-bit T1 fonts
\usepackage{hyperref}       % hyperlinks
\usepackage{url}            % simple URL typesetting
\usepackage{booktabs}       % professional-quality tables
\usepackage{amsfonts}       % blackboard math symbols
\usepackage{nicefrac}       % compact symbols for 1/2, etc.
\usepackage{microtype}      % microtypography
\usepackage{xcolor}         % colors
\usepackage{graphicx}
\usepackage{tcolorbox}
\usepackage{amsmath}
\usepackage{lipsum}
\usepackage{tabularx}
\usepackage{booktabs}
\usepackage{makecell}
\usepackage{multirow}
\usepackage{multicol}
\usepackage{bbding}
\usepackage{colortbl}
\usepackage{xcolor}
\usepackage{underscore}
\usepackage[normalem]{ulem}
\usepackage{latexsym}
\usepackage{float}
\usepackage{enumitem}
\usepackage{wrapfig}
\usepackage{caption}
\usepackage{footmisc}
\usepackage{amssymb}

\title{Baichuan Alignment Technical Report}

\author{
  Mingan Lin\textsuperscript{\rm 1}, 
  Fan Yang\textsuperscript{\rm 1}, 
  Yanjun Shen\textsuperscript{\rm 1}, 
  Haoze Sun\textsuperscript{\rm 1}, 
  Tianpeng Li\textsuperscript{\rm 1}, 
  Tao Zhang\textsuperscript{\rm 1}, 
  Chenzheng Zhu\textsuperscript{\rm 1} \\
 Tao Zhang\textsuperscript{\rm 1}, 
 Miao Zheng\textsuperscript{\rm 1}, 
 Xu Li\textsuperscript{\rm 1}, 
 Yijie Zhou\textsuperscript{\rm 1}, 
 Mingyang Chen\textsuperscript{\rm 1},
 Yanzhao Qin\textsuperscript{\rm 2},
 Youquan Li\textsuperscript{\rm 2} \\
 Hao Liang\textsuperscript{\rm 2},
 Fei Li\textsuperscript{\rm 1},
 Yadong Li\textsuperscript{\rm 1}, 
 Mang Wang\textsuperscript{\rm 1}, 
 Guosheng Dong\textsuperscript{\rm 1}, 
 Kun Fang\textsuperscript{\rm 1},
 Jianhua Xu\textsuperscript{\rm 1}\\
 Bin Cui\textsuperscript{\rm 2},
 Wentao Zhang\textsuperscript{\rm 2}, 
 Zenan Zhou\textsuperscript{\rm 1}\textsuperscript{$\clubsuit$},
 Weipeng Chen\textsuperscript{\rm 1}\\
 \\
 \textsuperscript{\rm 1}Baichuan Inc. ~~~~~~~~~~~~~~~~ \textsuperscript{\rm 2}Peking University \\
}

\begin{document}

\maketitle
% \renewcommand{\thefootnote}{$\diamondsuit$}
% \footnotetext[1]{Corresponding Authors}
\renewcommand{\thefootnote}{$\clubsuit$}
\footnotetext[1]{Project lead and Corresponding author: \texttt{zhouzenan@baichuan-inc.com}} 

\begin{abstract}
%In this report, 
We introduce Baichuan Alignment, a detailed analysis of the alignment techniques employed in the Baichuan series of models. This represents the industry's first comprehensive account of alignment methodologies, offering valuable insights for advancing AI research. We investigate the critical components that enhance model performance during the alignment process, including optimization methods, data strategies, capability enhancements, and evaluation processes. The process spans three key stages: Prompt Augmentation System~(PAS), Supervised Fine-Tuning~(SFT), and Preference Alignment. The problems encountered, the solutions applied, and the improvements made are thoroughly recorded.

\par Through comparisons across well-established benchmarks, we highlight the technological advancements enabled by Baichuan Alignment. Baichuan-Instruct is an internal model, while Qwen2-Nova-72B and Llama3-PBM-Nova-70B are instruct versions of the Qwen2-72B and Llama-3-70B base models, optimized through Baichuan Alignment. Baichuan-Instruct demonstrates significant improvements in core capabilities, with user experience gains ranging from 17\% to 28\%, and performs exceptionally well on specialized benchmarks. In open-source benchmark evaluations, both Qwen2-Nova-72B and Llama3-PBM-Nova-70B consistently outperform their respective official instruct versions across nearly all datasets. This report aims to clarify the key technologies behind the alignment process, fostering a deeper understanding within the community.
Llama3-PBM-Nova-70B model is available at~\url{https://huggingface.co/PKU-Baichuan-MLSystemLab/Llama3-PBM-Nova-70B}.

\end{abstract}

\begin{figure}[H]
    \centering
    \includegraphics[width=0.85\linewidth]{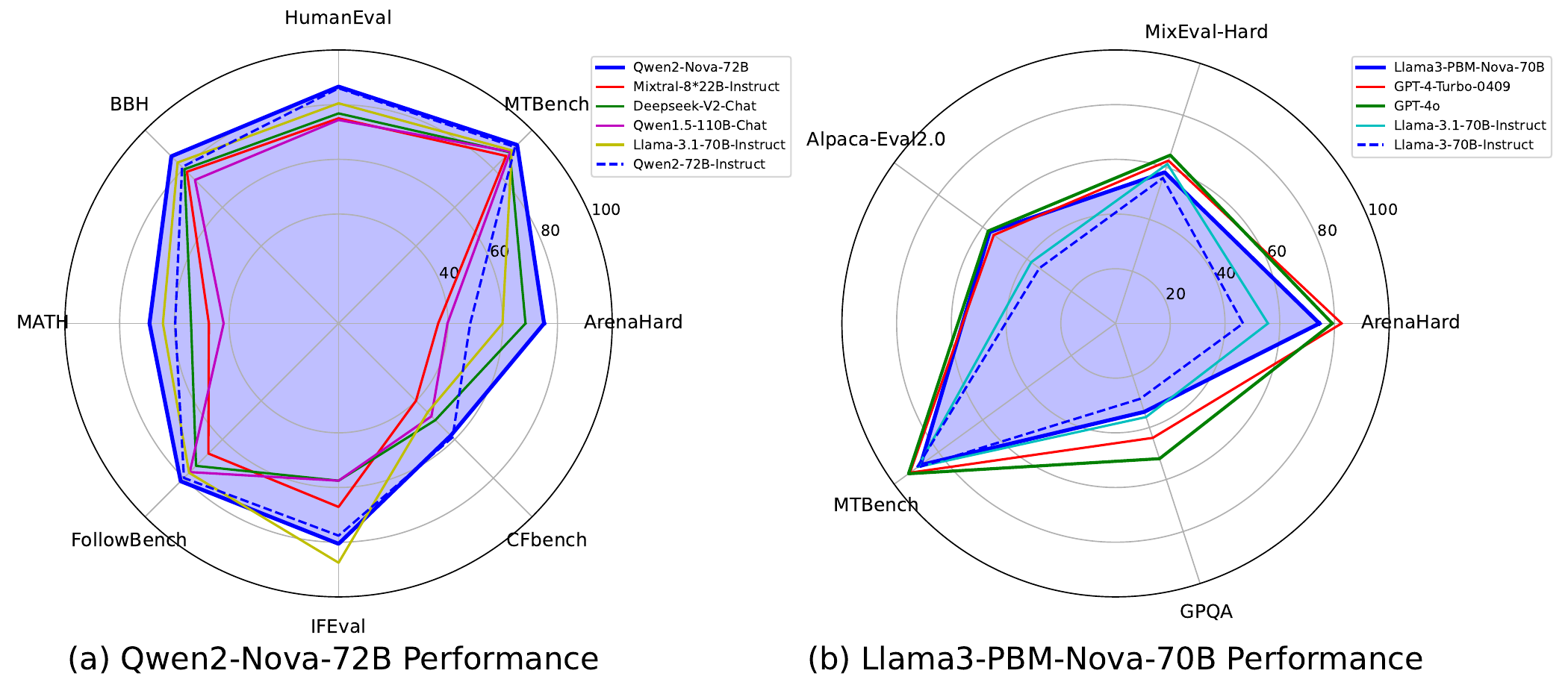}
    \caption{Performance Comparison of Qwen2-Nova-72B and Llama3-PBM-Nova-70B with Others}
    \label{ResultCompare}
\end{figure}

\iffalse
\begin{figure}[h]
    \centering
    \begin{minipage}[b]{0.5\textwidth}
        \centering
        \includegraphics[width=\textwidth]{picture/pic1_baichuaninstruct_nova_performance_radar.pdf}
        \caption{Instruction}
        \label{fig:half-width}
    \end{minipage}
    \hfill
    \begin{minipage}[b]{0.4\textwidth}
        \centering
        {and in point size 12. Two line spaces precede the abstract. The abstract, and in point size 12. Two line spaces precede the abstract. The abstract, and in point size 12. Two line spaces precede the abstract. The abstract}
    \end{minipage}
\end{figure}
\fi
\newpage
\tableofcontents

\newpage
\section{Introduction}  
% \ZT2{Wait until the V1 version is completed to summarize.}  

\iffalse
\begin{figure}[H]
    \centering
    \includegraphics[width=0.85\linewidth]{picture/introduction.png}
    \caption{The pipeline of baichuan alignment}
    \label{pipeline}
\end{figure}
\fi

In recent years, Large Language Models (LLMs) have achieved significant breakthroughs~\cite{radford2018improving,devlin2018bert,raffel2020exploring,brown2020language,chowdhery2023palm,achiam2023gpt,anil2023palm,anthropic2024claude,touvron2023llama,team2023gemini}, showing early signs of advancing toward Artificial General Intelligence (AGI). Through the mechanism of next-token prediction, LLMs undergo self-supervised pre-training on vast datasets, thereby acquiring a diverse spectrum of capabilities that empower them to perform an array of tasks, such as text continuation, summarization, and creative composition. Further advancements in alignment methodologies, such as Prompt Augmentation Systems~(\textbf{PAS})~\cite{zheng2024pas}, Supervised Fine-Tuning~(\textbf{SFT}), and preference modeling~\cite{ouyang2022training}, have become important to the evolution of LLMs. 
These developments have enhanced the models' ability to comprehend user intentions and adhere to instructions, thereby enhancing their conversational skills and adaptability to complex real-world scenarios. Despite its critical importance, a comprehensive understanding of alignment remains largely inaccessible to the broader public~\cite{bai2023qwen,yang2024qwen2,wang2024qwen2,chu2024qwen2,young2024yi,bi2024deepseek,liu2024deepseek,glm2024chatglm,hu2024minicpm,yang2023baichuan,touvron2023llama,achiam2023gpt}. The broad and intricate nature of alignment results in fragmented research efforts, which often provide only a narrow insight into specialized areas, making it challenging to present a holistic perspective of the alignment landscape. Additionally, alignment is frequently regarded as a proprietary cornerstone in the LLMs training, shrouded in corporate confidentiality. To foster advancement within the LLM community, we present a comprehensive and systematic exposition of Baichuan Alignment, featuring a suite of advanced and practical alignment techniques.

\par Baichuan Alignment comprises three critical phases: Prompt Augmentation Systems~(\textbf{PAS}), Supervised Fine-Tuning~(\textbf{SFT}) and Preference Alignment. The PAS stage aims to transform user queries into instructions that are more comprehensible and actionable for LLMs through automated Prompt Engineering (PE) techniques. The SFT stage equips LLMs with the ability to engage in dialogue and handle complex tasks using a large corpus of high-quality and diverse data. The Preference Alignment further aligns the LLMs with human values and preferences. 
This report primarily focuses on four aspects: optimization, data, key capability enhancement, and system evaluation, which are critical elements of Baichuan Alignment. The optimization ensures the effectiveness and efficiency of LLM training, and Section~\ref{Optimization} provides a detailed discussion on methods for training, prompt argumentation, and model merging, which collectively accelerate and improve model performance. In Section~\ref{Data}, we emphasize the importance of data in alignment, with a focus on prompt selection, response construction, and preference data. Section~\ref{subsec:key ability} outlines the challenges encountered in enhancing core capabilities on the path to AGI, detailing the specific approaches and insights gained through Baichuan Alignment. Finally, Section~\ref{subsec:eval} presents a comprehensive evaluation of Baichuan Alignment, focusing on user experience, open-source benchmarks, and specific capability assessments. This evaluation is crucial for assessing model capabilities and guiding iterative improvements, particularly through user evaluations and third-party product assessments that align closely with user experience. Additionally, we introduce tailored benchmarks, including CFBench~\cite{zhang2024cfbench}, SysBench~\cite{qin2024sysbench}, and FB-Bench~\cite{li2024fbbenchfinegrainedmultitaskbenchmark}, designed to address current challenges and practical needs in LLM applications.

We conducted a systematic evaluation of multiple models after Baichuan Alignment from various perspectives. Among these, Baichuan-Instruct serves as our internal model, while Qwen2-Nova-72B and Llama3-PBM-Nova-70B are instruct versions derived from the Qwen2-72B~\cite{yang2024qwen2} and Llama-3-70B~\cite{dubey2024llama} base models, respectively, fine-tuned using Baichuan Alignment techniques. User experience assessments reveal that Baichuan-Instruct exhibits significant improvements across several core capabilities, with enhancements ranging from 17\% to 28\%. Notably, there are marked improvements in mathematics and reasoning, with increases of 28\% and 23\%, respectively. In terms of open-source benchmarks, Qwen2-Nova-72B demonstrates substantial advancements over its official instruct version, Qwen2-72B-Instruct, across multiple leaderboards. Similarly, Llama3-PBM-Nova-70B shows significant improvements compared to Llama-3-70B-Instruct, particularly achieving a 60\% relative increase (from 46.6 to 74.5) on the ArenaHard benchmark. On specific benchmarks, Baichuan-Instruct ranks competitively in core competencies on CFBench, SysBench, and FBBench, compared to the leading models currently available. This multidimensional, comprehensive, and in-depth evaluation underscores the superiority of Baichuan Alignment technology and its applicability across different foundational models. As a crucial step toward AGI, we are publicly sharing the challenges encountered during the Baichuan Alignment process, the solutions devised, and some in-depth insights. This disclosure aims to provide the community with new perspectives or lessons learned from failures, thereby fostering discussions on alignment and making a meaningful contribution to the advancement toward AGI.

\section{Optimization}\label{Optimization}

\subsection{Training}\label{sec:trainng}

\paragraph{SFT}

During supervised fine-tuning, we optimize models using a learning rate of $1 \times 10^{-5}$ and conduct training over 2 to 6 epochs for models with various size. For all training sessions, we employ sample packing, which will be discussed below. Additionally, we apply weight decay to prevent overfitting.

\paragraph{Reward}
In many practice \cite{ouyang2022traininglanguagemodelsfollow, bai2023qwentechnicalreport, liu2024deepseek}, a reward function is estimated based on the preference data using the Bradley-Terry \cite{19ff28b9-64f9-3656-ba40-08326a05748e} model. However, the Bradley-Terry model has some limitations, including a tendency to overfit the data. This can occur even before completing a single epoch of training despite having a high-quality datasets. Besides, the reward model only ensure the relative order of the reward score for different responses, not their absolute feeling. Eg: the fitted reward score difference of $S_{perfect} - S_{bad1}$ may less than $S_{bad1} - S_{bad2}$, which encourage the model to find a shortcut to hack the reward. To mitigate such impacts, we add a point wise MSE loss to let the model fit the normalized absolute score which annotated from Section~\ref{subsec:pref-data}. Thus the reward model will minimize the following objective:
\begin{equation}
\begin{split}
        \mathcal{L}_{\theta} = E_{(x, y_w, y_l)} \Big[ & -\log \left(\sigma(r_\theta(x, y_w) - r_\theta(x, y_l)) \right) \\
        & + \alpha \left((r_\theta(x, y_w) - \hat{r}_x^{y_w})^2 + (r_\theta(x, y_l) - \hat{r}_x^{y_l})^2 \right) \Big] 
\end{split}
\end{equation}
where the $\alpha$ is a adjustable coefficient, $\hat{r}_x^{y_w}$ and $\hat{r}_x^{y_l}$ are the normalized annotated absolute score for the chosen and rejected data. Note that in our preference dataset, we also mixed some dataset without absolute score, like open source hh-rlhf\cite{bai2022traininghelpfulharmlessassistant} dataset, SHP\cite{pmlr-v162-ethayarajh22a} dataset. For data in these dataset, the $\alpha$ is set to 0.
With this objective, we find the fitted reward model is more robust. The RM model is trained for 1 to 2 epochs, depending on its size.

\paragraph{Reinforcement Learning}
During the course of reinforcement learning, we conducted experiments on both PPO\cite{schulman2017proximalpolicyoptimizationalgorithms} and GRPO \cite{shao2024deepseekmathpushinglimitsmathematical} to further enhance our model. When applying these two approaches, we made few modifications: 
When doing the Reinforcement Learning to merged models(described in section \ref{subsec:model-merge}), using cross-entropy loss on SFT dataset as PTX loss would degrade the model into an SFT+reinforcement model. 
Therefore, during the reinforcement training, we use the KL divergence between the policy model and the original model as PTX loss. Furthermore, when calculating KL divergence of each token, we first select the indices corresponding to the Top500 logits from the reference model. For these Top500 indices, we separately computed the normalized log probabilities from both the policy model and the reference model. We then applied the standard KL divergence formula between the two sets of log probabilities, rather than using the simplified KL divergence version \cite{schulman2020approximatingkl}, to ensure the KL remained non-negative and as accurate as possible. In GRPO, the n\_sample for each prompt is set to 3 and we did not remove the KL term in token-level reward decomposition. The optimization objective of the Reinforcement Learning would described as below:  
% $$ objective(\theta) = \mathbb{E}_{(x, y) \sim D_{\pi_{\theta}^{RL}}} \left[ r_{\theta}(x, y) - \beta \mathbb{KL}\left( \pi_{\theta}(y|x) ||\pi_{ref} (y|x) \right) \right] + \gamma \mathbb{KL}_{x \sim D_{pretrain}} \left[\pi_{\theta}(x)||\pi_{ref}(x)\right]
% $$
\begin{equation}
\begin{split}
        \text{objective}(\theta) = & \, \mathbb{E}_{(x, y) \sim D_{\pi_{\theta}^{RL}}} \left[ r_{\theta}(x, y) - \beta \, \mathbb{KL}\left( \pi_{\theta}(y|x) \, || \, \pi_{ref} (y|x) \right) \right] \\
        & + \gamma \, \mathbb{KL}_{(x, y) \sim D_{SFT}} \left[ \pi_{\theta}(y|x) \, || \, \pi_{ref}(y|x) \right]
\end{split}
\end{equation}
During the training process, the model would be save and evaluated after every certain training steps. We establish corresponding test sets for various task categories to monitor the over-fitting or under-fitting of each category. 
This allows us to adjust the prompts proportion for each category in the training data, thereby enhancing the overall performance of the model.

In our experiment, we found that even without the critic model, GRPO can still achieve comparable results with PPO. The evaluation benchmark(described in Section~\ref{subsubsec:benchmarks}) difference between GRPO and PPO $\Delta_{GRPO-PPO}$ is $+3.7\%$ for FollowBench, $+1.0\%$ for CFBench PSR Full and $-0.50\%$ for SysBench. On the other hand, GRPO can save nearly half of the training resources compared to PPO. Additionally, it achieves better performance than direct optimization methods like DPO\cite{rafailov2024directpreferenceoptimizationlanguage} and KTO\cite{ethayarajh2024ktomodelalignmentprospect}, while requiring only a marginal increase in training resources. Therefore, the GRPO is chosen as our Reinforcement Learning method.

\subsection{Efficient Training}\label{subsec:eff-train}

\paragraph{Packing} To address training inefficiencies stemming from the varying lengths of samples, we consolidated multiple short samples into a single long sample, significantly reducing the number of padding tokens required \cite{hermes3}. Traditional packing is performed at the sample level and typically implemented on the \texttt{attention_map} of Flash Attention v1. In our approach, we leveraged the \textit{cu_seqlens} argument in Flash Attention v2 \cite{FlashAttention2}, which allows for mask-free variable sequence lengths. This feature effectively isolates attention across different samples, preventing contextual contamination between them and enabling plug-and-play equivalent training. Our experiments demonstrated that this method of packing increased the effective token utilization rate within a batch from 10\% to 98\%, achieving nearly a tenfold improvement in efficiency without any loss in performance.

\begin{figure}[htbp]
    \centering
    \includegraphics[width=\linewidth]{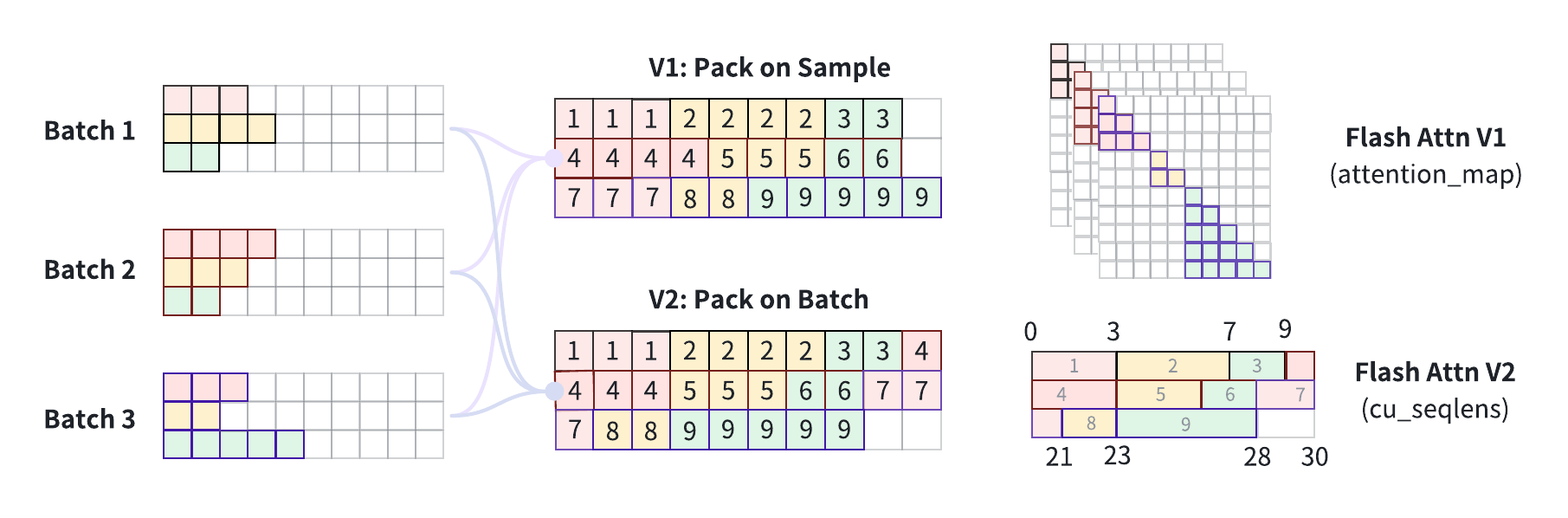}
    \caption{Difference between packing on sample and pakcing on batch.}
    \label{fig:pack}
\end{figure}

\paragraph{Multi-layer Gradient Checkpointing}
Conventional practical implementation of gradient checkpointing typically involves setting a checkpoint for each decoding layer for LLMs \cite{gradient-ckpt}.
However, in models with a very large number of layers, such as those with over 80 layers and more than 70 billion parameters, this setup is far from the most memory-efficient configuration. 
Gradient checkpointing is essentially a trade-off between time and memory. Ideally, the configuration should allow the GPU memory to accommodate a sequence length that fully utilizes the computational power, making it the most cost-effective setup. In practice, this parameter requires experimental tuning.
Therefore, we use the multi-layer gradient checkpointing that merges multiple decoding layers to achieve better memory control.
Assuming that during the forward pass of the model, each decoder layer has $n$ activations, then during the backward pass, the amount of storage required for gradient checkpointing is $hidden\_size \times n + hidden\_size \times num\_layer$. 
If $k$ layers are merged before checkpointing, the required storage becomes $k \times hidden\_size \times n + hidden\_size \times \frac{num\_layer}{k}$. This implies that when $k = \sqrt{\frac{num\_layer}{n}}$, the memory usage is minimized. It is important to note that the value of $n$ and the calculation of the optimal $k$ may vary depending on the specific architecture.
In our practice, this optimization can reduce the minimum number of GPUs required to train a model more than 70 billion with a sequence length of 16K from 128 GPUs to 40 GPUs.

\paragraph{Sequence Parallel}

Following DeepSpeed-Ulysses \cite{Ulysses}, we adapt the sequence parallelism method to our training framework, and achieve efficient and scalable training for large language models with extremely long sequence lengths. This makes it more suitable for high-frequency experiments and scenarios with limited total resources.

\subsection{Prompt Augmentation}
Variations in application contexts demand tailored response paradigms from large-scale language models.
For professional inquiries, the model should adopt an authoritative and informative tone, while for consumer interactions, it should strive for an empathetic and engaging tone. 
In both cases, the model's ability to adapt its style to the context is crucial. The model's flexibility in adjusting its communication style is key to meeting the diverse needs of different user groups.
Nonetheless, inconsistent response styles in training data may adversely affect the model's performance. If the data it is trained on lacks consistency, the model may struggle to develop a coherent and effective communication strategy. A feasible approach is to decouple the model's capabilities from the requirements for its response style. This allows for greater flexibility in optimizing and fine-tuning the model's functionality, while also enabling the customization of its response style according to   
varied needs or scenarios.

The effectiveness of large language models in various applications largely depends on the prompts' quality. 
There are already many designed prompts that can significantly enhance the performance of LLMs \cite{kojima2022large, wang2022self, wei2022chain, yao2024tree}. However, these methods that rely on manual prompt engineering are far less scalable and have steep learning curves and significant time investment for consumer-side users. Therefore, it is essential to explore the development of mechanisms for automatic prompt engineering, which could significantly enhance the efficiency and effectiveness of AI interactions.

We designed a plug-and-play system to automatically generate prompt complementary, named Prompt Augmentation System (PAS). The pipeline, in general, generates corresponding complementary content based on the user's prompt, and then the prompt is supplemented with the content and input into the main model. PAS usually supplements the following aspects of content: 1) Requirements for product applications scenario responses. 2) The extension content based on user intent with non-mandatory tunes. 3) Response format constraints, like "the content should be logically clear, organized, and easy for users to understand."

\begin{figure}[htbp]
    \centering
    \includegraphics[width=\linewidth]{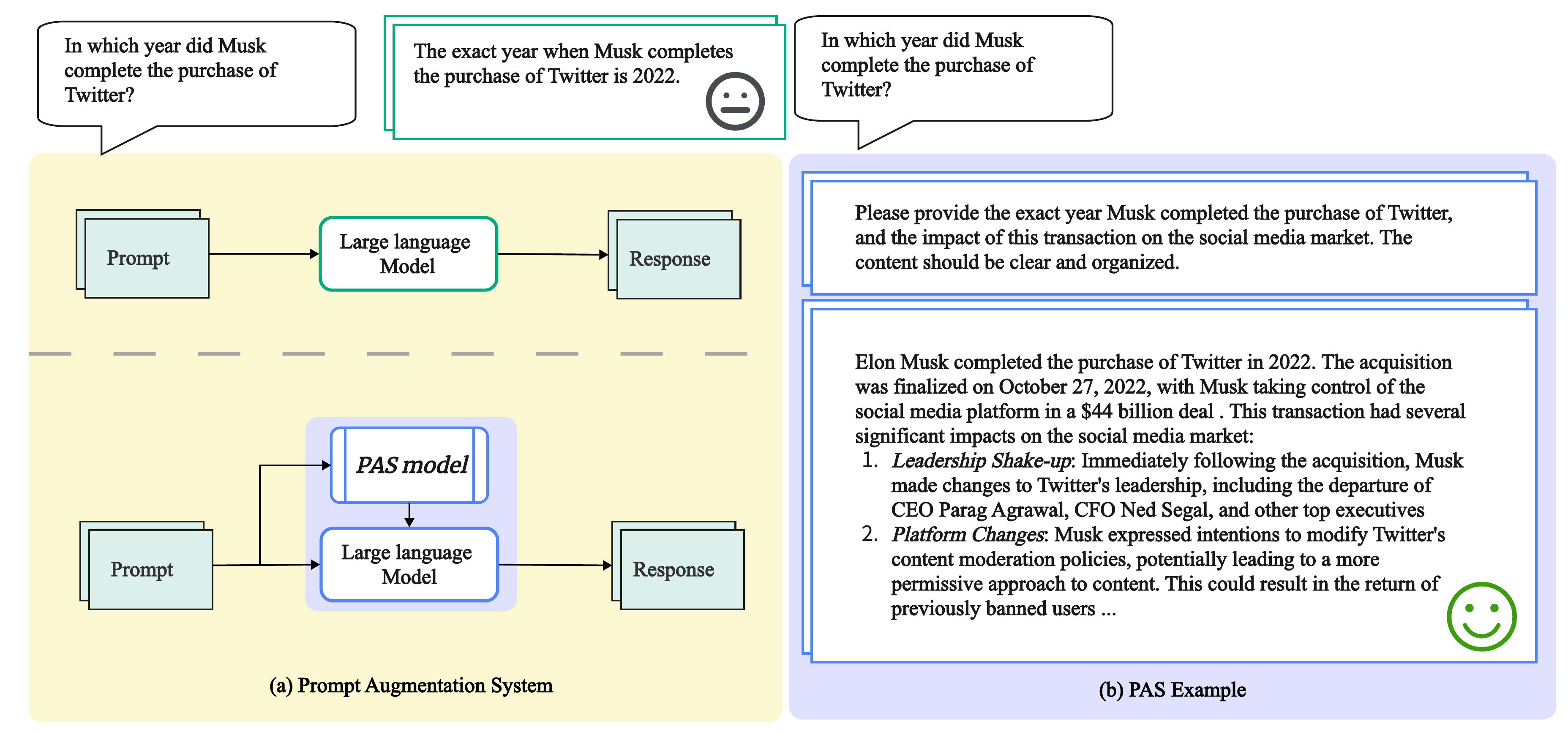}
    \caption{We present Prompt Augmentation System (PAS). (a) It takes user prompts, enhances them, and inputs the augmented prompts into LLMs. (b) PAS significantly improves responses across all categories in human evaluation.}
    \label{fig:pack}
\end{figure}

Regarding Retrieval Augmented Generation (RAG) for LLMs, the extended prompt content needs to be within the scope of the search results; therefore, when generating supplementary prompts, the search keywords should be used as new constraints.

\subsection{Model Merging}\label{subsec:model-merge}
Model merging is an emerging technique in the field of artificial intelligence that involves combining the parameters of multiple models, each optimized for different tasks, to create a more flexible and universal model. Despite being a relatively nascent area of research, model merging is rapidly advancing and has already demonstrated its utility across various domains. A notable application of this technique is in the enhancement of foundation models. By merging models fine-tuned on different downstream tasks, the capabilities of large language models can be significantly augmented. For an in-depth exploration of model merging, we refer readers to a comprehensive survey of merging algorithms~\cite{yang2024model}.

One of the challenges in model training is the proliferation of fine-tuned checkpoints, which result from different training data updates, hyperparameter settings, and regularization techniques. These variations often lead to divergent performance outcomes across different domains, a phenomenon commonly referred to as the "seesaw effect", where improvements in one domain result in deteriorations in another. Model merging offers a promising solution to mitigate this effect by balancing the performance across diverse tasks.

In practice, we selected the best-performing models from different domains and applied various merging algorithms, including Linear, Task Arithmetic~\cite{ilharco2022editing}, and Model Stock~\cite{jang2024model}, using the model merging toolkit, MergeKit~\cite{goddard2024arcee}. Our experimental results indicate that the merged models typically achieved more balanced performance across the evaluated domains. Among the tested algorithms, Model Stock consistently delivered the best overall performance.

\section{Data}\label{Data}
\begin{figure}[th]
    \centering
    \includegraphics[width=0.95\linewidth]{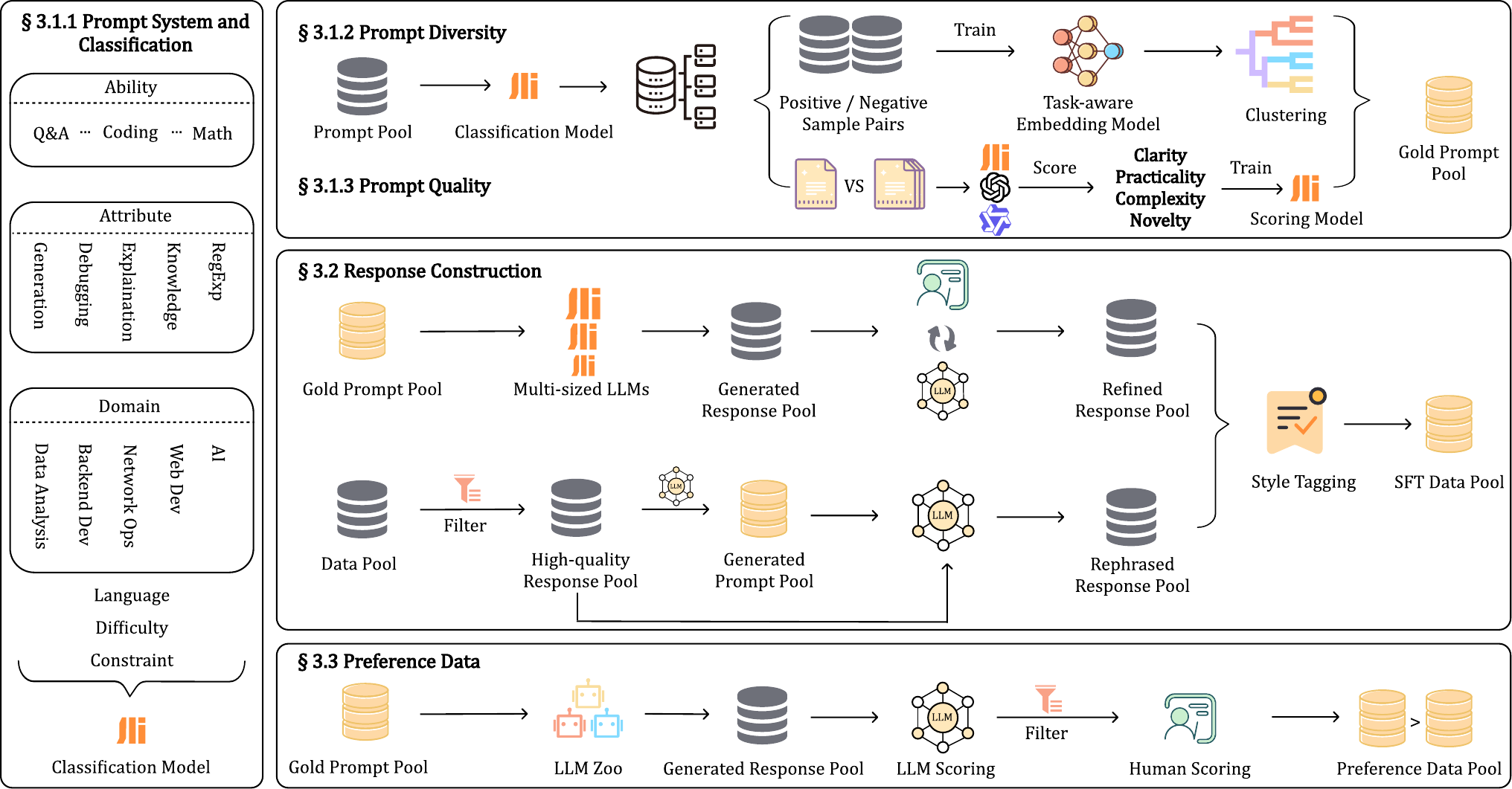}
    \caption{
The Pipeline of Alignment Data Processes, including: Prompt System and Classification, Prompt Selection, and Construction of Response and Preference Data}
    \label{fig:Pipeline of Data Processes}
\end{figure}

Alignment data has been extensively validated as critical to the ultimate performance of LLMs, includeing both prompt response pairs and preference data~\cite{li2024quantity,xia2024less,kung2023active,du2023mods,ge2024clustering}. Figure~\ref{fig:Pipeline of Data Processes} delineates the comprehensive pipeline for constructing the Baichuan alignment dataset. Initially, we develop an prompt system and a classification model (Section \ref{subsec:Prompt System and Classification}), which form the foundational basis of the data flywheel. We then elaborate on the pivotal steps in data construction, focusing on prompt diversity (Section \ref{subsec:Prompt Diversity}) and prompt quality (Section \ref{subsec:Prompt Quality}). Furthermore, we detail various techniques and best practices in generating responses (Section \ref{subsec:Response Construction}). Additionally, we provide insights derived from the process of constructing preference data (Section \ref{subsec:pref-data}).

\subsection{Prompt Selection}\label{subsec:data-pipeline}
\subsubsection{Prompt System and Classification}\label{subsec:Prompt System and Classification}
The prompt system plays a essential role in data management, augmenting the breadth and balance of data coverage, and steering the reliability and direction of model performance iterations. Guided by taxonomy and statistical analysis, and supplemented by human expertise, we have developed a multi-dimensional, multi-granularity prompt labeling system through numerous iterations and refinements. Subsequently, an automated labeling classification model is trained and deployed across various facets of data production.

\paragraph{Prompt System} We curated a diverse initial prompt pool, incorporating real requests from human prompts and various open-source datasets. Leveraging a general model, preliminary labels were assigned to these prompt. Subsequently, an initial prompt labeling system was developed through techniques such as synonym merging, long-tail pruning, and hierarchical association. Through iterative refinements and expansions by human annotators, the final version of the labeling system and the enriched label dataset were meticulously established.    

Notably, our  system is meticulously structured around six primary dimensions: ability, attributes, domains, language, difficulty and prompt constraints. These dimensions, featuring both interwoven and hierarchical structures, enable the creation of thousands of combinatory types. Ability encompass the skills required by the LLM to complete tasks, including knowledge-based Q\&A, text generation, code programming, and logical reasoning. Attributes provide unique contextual information linked to capabilities, such as literary writing and practical writing for text generation. Domains connect the real world to the LLM, imbuing it with relevance and vitality, with common domains including IT, history, and humanities. Language defines the medium of expression, categorized into Chinese, foreign languages, and programming languages, with further subdivisions such as Simplified Chinese, Traditional Chinese, Classical Chinese, English, French, Python, and C++. Difficulty indicates the complexity of prompt, classified into easy, intermediate, and difficult levels. Prompts constraints highlight the importance of constraints within prompts, categorized by the number of constraints into unconstrained, simple constraints, and strong constraints. This integrated framework ensures a comprehensive and adaptable system, adept at addressing a wide array of instructional scenarios.

\paragraph{Classification Model} Under the guidance of the aforementioned prompt system, we initially utilized advanced LLMs to classify prompts. This process was refined through methods such as voting and manual verification of low-confidence samples, resulting in the construction of a training set comprising tens of thousands of examples. Subsequently, we fine-tuned a specialized automatic prompt labeling model based on Baichuan2-13B~\cite{yang2023baichuan} and the label data, achieving a 90\% accuracy rate on the evaluation set through prompt-based techniques. In contrast, a fine-tuned BERT~\cite{devlin2019bert} model only reached an accuracy of 81\%. The integration of the prompt system with the automatic classification model enables efficient management and iteration of the prompt set. This includes evaluating prompt diversity and coverage, grouping for mining and optimization, and automatically matching prompts to specialized human annotators. Every new data point added to the training set undergoes this systematic process, ensuring consistent and comprehensive handling of all prompt-related tasks, as shown on the far left of Figure \ref{fig:Pipeline of Data Processes}.

\subsubsection{Prompt Diversity}\label{subsec:Prompt Diversity}
Numerous studies have demonstrated that similar and repetitive prompts can adversely affect model performance, underscoring the critical importance of prompt diversity during the alignment phase~\cite{wei2021finetuned,wu2023self,chen2023maybe}. In practical applications, prompts typically comprise both instructions and inputs, often formatted as task templates concatenated with contextual information. Current semantic-based dense embedding methods generally model the overall semantics but fail to adequately capture the repetitive representation of task-specific information~\cite{reimers2019sentence,xiao2024c}. To address this challenge, we propose a task-aware embedding model that more precisely captures the nuanced differences between instructions, thereby facilitating the selection of a more diverse set of prompts.

As depicted in the upper right section of Figure \ref{fig:Pipeline of Data Processes}, our primary innovation lies in extracting high-quality task-aware training data through a multi-granularity clustering approach. We begin with coarse-grained clustering, followed by fine-grained clustering within each category. Using the Longest Common Subsequence (LCS) algorithm and heuristic rules, from different fine-grained clusters, samples with similar task templates are identified as hard positive samples, while those without similar task templates are identified as hard negative samples. These samples are then used in contrastive learning with Triplet Loss to train a high-quality embedding model. Additionally, we incorporate hierarchical clustering principles by setting an incremental sequence of thresholds to perform layered clustering, thereby enhancing the robustness, efficiency, and stability of the algorithm. This combined approach allows us to achieve significantly superior results compared to the original algorithm, using only 50\% of the original data volume.

\subsubsection{Prompt Quality}\label{subsec:Prompt Quality}

High-quality prompts are instrumental in training models more efficiently to achieve superior performance~\cite{zhou2024lima,xu2023rethinking,liu2023makes}. However, existing prompt quality scoring methodologies exhibit certain deficiencies, as they either lack robustness across general datasets or fail to satisfy the personalized instruction screening requirements in specific contexts~\cite{he2024shed,lu2023instag,li2023one,caoinstruction,chen2023alpagasus}. To address these issues, we have developed a flexible and scalable automated prompt quality evaluation framework leveraging large language models (LLMs).

\paragraph{Training Data}Drawing inspiration from the ArenaHard evaluation system~\cite{li2024crowdsourced}, we utilize a pairwise LLM-based judging mechanism to construct training dataset. Specifically, we initially employ the classification model outlined in Section~\ref{subsec:Prompt System and Classification} to categorize instructions into 20 distinct buckets. From each bucket, we randomly select 30 data points to serve as anchors. Subsequently, other data within the bucket are paired with these anchors, and  multiple LLMs are employed as the evaluator. The evaluation process considers four dimensions: Clarity, Practicality, Complexity, and Novelty, using a three levels scoring system for pairwise comparisons. The aggregated scoring results from multiple anchors are ultimately used to assign the final quality label to each example.

\paragraph{Performance and Effectiveness}In consideration of performance and efficiency, we fine-tuned the Baichuan2-7B~\cite{yang2023baichuan} base model using the aforementioned training data to develop the final prompt scoring model, Quality-7B. This model offers substantial advantages in terms of efficiency and cost compared to powerful LLMs such as GPT-4~\cite{achiam2023gpt}. Furthermore, through testing on a set of 200 evaluation samples, we have demonstrated that the Quality-7B model's scoring accuracy significantly surpasses GPT-4.

\subsection{Response Construction}\label{subsec:Response Construction}
\paragraph{Human Annotation}\label{subsec:human-anno}
We employ various scales of models and diverse generation strategies to sample multiple responses for the same prompt. These responses are automatically assigned to specialized annotators based on type labels for preference ranking, which greatly enhances annotation efficiency and quality, thereby significantly elevating the upper limit of data. Typically, we use a Reward Model or LLM-as-Judge to pre-filter responses, forming a response set with graded quality distinctions to collect a sufficiently rich preference order. When the best response fails to meet the established standards, we require annotators to modify the answers, thereby constructing a high-quality SFT and preference dataset.

\paragraph{Human-Machine Collaborative Annotation}
Large-scale annotation demonstration datasets are constrained by cost, time, and personnel. Therefore, we adopt a human-machine collaborative approach to enhance efficiency. For example, we use the critique from LLM as auxiliary information to improve the speed and quality of response modifications. Sometimes, we ask evaluators to compile a set of quality defects based on a sampled data subset, and then use LLM to perform defect mining and automatic rewriting. In practice, many useful techniques have been accumulated during the annotation process, and any attempts or methods that utilize LLM to enhance annotation efficiency are highly rewarded.

\paragraph{Instruction Back-Translation}
For tasks requiring expertise or creativity, it is unrealistic and uneconomical to expect annotators to always produce high-quality responses. To address this issue, inspired by~\cite{li2023self}, we use text quality models to filter high-quality texts from public sources across various fields and synthesize high-quality prompt-response pairs. We have collected a pool of high-quality copywriting, exemplary essays, and highly praised posts, and then back-translate them to generate corresponding creative instructions.

\paragraph{Personalization}\label{subsec:Personalization}
In the context of multiple correct responses to the same prompt, individuals may exhibit divergent preferences: some favor concise and direct answers, while others prefer detailed and structured responses. By incorporating style descriptions corresponding to different response styles into the prompts, we mitigate style preference conflicts and enhance the model's adaptability to switch between styles. Additionally, responses often include refusals due to constraints related to values and hallucinations. Prompts concerning safety, timeliness, and model functionality can easily lead to excessive refusals, significantly reducing usability and harming user experience. We address this issue by strictly controlling the proportion of refusals and incorporating refusal constraints into the system message and prompt.

\subsection{Preference Data}\label{subsec:pref-data}
A dataset with high data quality and good data diversity is important not only for SFT dataset, but also for RLHF preference dataset. 
A data filter pipeline similar to Section~\ref{subsec:data-pipeline} is performed to get the high quality and high diversity prompt collections with their labels for the preference dataset. 
In addition, we filter prompts with simple, time-sensitive meaningless or out-of-ability tasks using the classified labels. We also only keep the prompts with Chinese or English languages due to our annotators language limitations.

For the data annotation, we refer and improve the process pipeline of Llama 2 \cite{touvron2023llama2openfoundation}. 
We use our Top 3 most advanced models to sample 5 responses for each prompt.
The generation setup of each model is $Temperature=1$, $TopP = 0.99$, $TopK = 50$. To further increase the data diversity and filter the prompts which model already answers well, a rouge\cite{linchinyew2004rougescore} rule based filter is used to roughly filter the similar responses. For different set of tasks, we developed several LLM-as-Judge\cite{zheng2023judgingllmasajudgemtbenchchatbot} prompts named AutoRator to evaluate the judged score of the response. In summary, there are three types of the AutoRator used in the judge pipeline: 1) Absolute score: Judge the absolute evaluate score for tasks with open responses like Writing, Open QA and etc. 2) Pair comparison: Judge if the two response have the similar thought chain and same result. For tasks which have specific answers like Math, Reasoning. 3) Golden Answer: Judge the absolute evaluate score with respect to the golden answer for datasets which have golden answers. After the AutoRator evaluation, prompts with all high-scoring responses or very similar scoring responses are removed, and the remaining data is sent to annotators. For some math and reasoning data with golden answers, we find the LLM-as-Judge score is good enough and directly use them. 

When doing the annotation, for each group of response, we not only ask for annotators to rate the their preferred order, but also the absolute feeling score about the helpfulness, writing fluency and safety. To ensure a good annotation quality and for possible future use, we ask the annotators to highlight the erroneous part and add a note about the error if a response has factual errors or logical errors. If all of responses of a prompts get low absolute annotation score, it would be send to the response modify process described in Section~\ref{subsec:human-anno}.

Due to the majority of our annotated data is in Chinese language, we have incorporated additional open-source datasets in other languages to enhance the multilingual capabilities. These datasets include hh-rlhf\cite{bai2022traininghelpfulharmlessassistant}, Helpsteer2\cite{wang2024helpsteer2}, UltraFeedback\cite{cui2023ultrafeedback} and SHP\cite{pmlr-v162-ethayarajh22a}.

\section{Key Ability}\label{subsec:key ability}

\subsection{Instruction Following}\label{subsec:Instruction Following}
\begin{figure}[th]
    \centering    \includegraphics[width=1\linewidth, trim=85 50 85 22, clip]{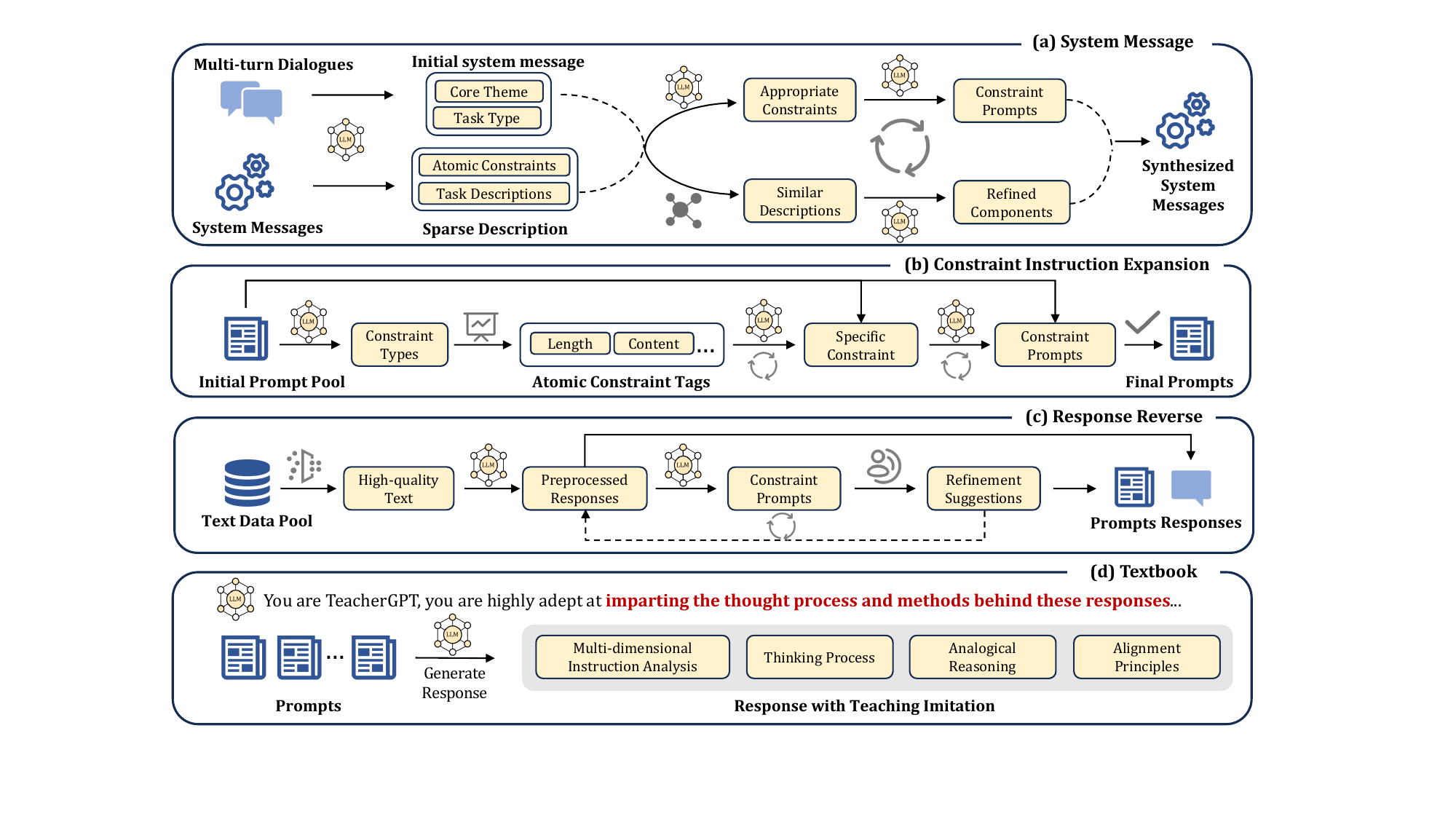}
    \caption{Overview of Instruction-Following Optimization, including: System Message, Constraint Expansion, Response Reversal, and Textbook Techniques.}
    \label{fig:Overview of Instruction-Follow}
\end{figure}

The ability to following instruction is a critical capability of any advanced Large Language Model (LLM), particularly when confronted with the intricacies of real-world applications. Baichuan Alignment has implemented several key enhancements to bolster this crucial capability, as depicted in Figure~\ref{fig:Overview of Instruction-Follow}. These primarily include the construction of complex system messages, expansion of instruction constraints,
response reversal, and textbook learning.

\paragraph{System Message}\label{System Message}
System message is a set of special instructions located at the beginning of multi-turn conversations, pre-setting the role, background, approach or output format of the model to align with the user-defined objectives~\cite{wallace2024instruction,lee2024aligning,lu2024sofa,mu2023can}. 
We enhance the understanding and adherence capabilities of LLMs by collecting and constructing a large-scale, high-quality dataset of system messages (see Figure~\ref{fig:Overview of Instruction-Follow}~(a) for details).
Initially, we collect a substantial amount of preliminary system message components. On one hand, we gather direct system messages from real user logs and open-source data, further decomposing them into task descriptions, workflows, output specifications, constraints, and initial statements. On the other hand, we extract samples with complex constraints and requirements from extensive multi-turn dialogues, utilizing LLMs to extract and synthesize similar components. Subsequently, we employ clustering and filtering techniques to obtain a diverse array of constraint descriptions and other component pools. We seamlessly integrate the original prompts with appropriate constraints to create constraint-rich prompts. Simultaneously, we meticulously select the remaining components of the system messages and further embed the constraint prompts into suitable positions, accompanied by evolutionary learning and constraint rationalization steps. This process is iterated multiple times, culminating in a final filtering step to eliminate unreasonable system messages, resulting in a comprehensive system prompt database. Through these steps, we ensure that the system messages cover a diverse range of task types, such as role-playing, function call, and security. We also ensure the tight integration of constraint descriptions and task descriptions, as well as the overall high quality of the data.

\paragraph{Constraint Instruction Expansion}\label{Constraint Instruction Expansion}
The construction of complex instructions through the addition of constraints has been well-documented as crucial for enhancing instruction-following capabilities~\cite{sun2024conifer,dong2024self,he2024complex,xu2023wizardlm}. To ensure the quality and coverage of constraint-based instruction data, we employ a series of steps and methodologies, including the development of a complex constraint system, collaborative iteration with multiple LLMs, and independent validation, as illustrated in Figure~\ref{fig:Overview of Instruction-Follow}~(b). Drawing on previous work\cite{zhang2024cfbench}, we utilize LLMs to decompose, cluster, and synthesize over 30 types of constraints from a large pool of prompts. Through sampling, we verify that the coverage of constraint types in random instructions reaches up to 90\%, and obtain rich descriptions corresponding to each type of constraint. Subsequently, we employ multiple LLMs to increase the complexity of constraints in selected original prompts. This process involves detailed prompt engineering guidance to ensure LLMs select constraint descriptions that match the prompts, focusing on the rationality of constraint types, the flexibility of constraint expression, and filtering based on prompt quality scores. Finally, we use a separate LLM as a validator to select high-quality, constraint-rich instructions.

\paragraph{Response Reversal}
The alignment of complex instructions poses two key challenges: (1) the scarcity and lack of diversity in high-quality prompts containing multiple constraints, and (2) the difficulty and high cost associated with producing responses that adhere precisely to these prompts. Nevertheless, there is an abundance of high-quality natural language text that can be utilized by leveraging the inherent attributes of these texts as constraints for generating corresponding instructions. This approach facilitates the large-scale creation of rich and high-quality prompt-response pairs with multiple constraints. To address this, we have developed a response reversal pipeline, as illustrated in Figure~\ref{fig:Overview of Instruction-Follow}~(c). The detailed process is as follows: First, we perform quality filtering by employing a quality scoring model to extract high-quality text from knowledge-rich passages in web text. Next, useing this high-quality text as the response and tasks a LLM with generating the corresponding prompt, incorporating the inherent attributes of the response as constraints for the prompt. Finally, the LLM inspector evaluates the quality of the generated prompts, assigns scores, and provides suggestions for improvement. This process is iterated until all constraints present in the response are fully encapsulated within the prompt.

\paragraph{Textbook}\label{Textbook}
While simple imitation learning can efficiently acquire relatively straightforward instructions, it often results in only a superficial understanding when applied to complex instructions. To address the issue where models may know the outcome but lack comprehension of the underlying reasoning, we propose a method of instructional imitation. By explicitly focusing on learning the intent of the instruction, the thought process, analogical reasoning, and alignment principles, we can significantly enhance the efficiency of instruction learning and improve the model's generalization and alignment capabilities. Specifically, for complex instructions, we leverage LLMs to generate responses that are highly aligned with the prompts, as illustrated in Figure~\ref{fig:Overview of Instruction-Follow}~(d). For example, when tasked with understanding an instruction, the model might be prompted to "explain how to comprehend the instruction, analyzing its details and challenges from multiple dimensions such as professionalism, complexity, and output constraints." Alternatively, LLMs can be guided to articulate the thought process behind a solution, such as "elaborate on the reasoning process behind the solution, including relevant domain knowledge, reasoning methods, and common pitfalls, particularly focusing on erroneous reasoning paths and easily overlooked output constraints like format, style, tone, length, and word count limitations." Additionally, prompts like "you need to learn the reasoning techniques and background knowledge mining from the given examples. For instance: xx" can be fed to the LLM to endow it with the ability to generalize and adapt to different task scenarios. Some alignment principles are also taught, such as "avoid using technical terms or complex expressions whenever possible" and "consider the user's emotional state," which are common response techniques. This approach ensures a deeper understanding and more robust application of complex instructions, thereby enhancing the overall capability of the model to generalize and align with diverse instructional requirements.

\subsection{Math}\label{subsec:ability math}

\paragraph{Prompt Collection} During the prompt selection phase, a multi-level balanced approach is employed to maintain problem diversity. Specifically, problems are sampled evenly across elementary, middle, high school, and college levels. Furthermore, we use more fine-grained criteria, called ``knowledge points'', where within each level, problems are selected to evenly cover over 1,000 different knowledge points. Priority is given to sampling problem-solving questions, as they generally include more comprehensive steps.

\paragraph{Response Generation} In many datasets of math problem-solving, the provided answers often lack detailed steps. Manual annotation is a common method, however, some intermediate steps may be ignored. Nevertheless, these steps are important for model training.
By leveraging large language models in combination with prompts and standard answers, we can generate detailed solutions that maintain a consistent style based on the reference answers. This approach synthesizes a substantial amount of math alignment data with detailed steps, significantly enhancing the model's mathematical capabilities. Directly generated answers often struggle to match correct answers, so providing reference answers and steps helps the model enrich these into coherent reasoning chains. We find that reference answers significantly improve the quality of generated responses. However, since reference answers may contain errors, we instruct the model to consider them as having a 95\% probability of correctness. This approach reduces incorrect inferences based on potentially faulty references. Furthermore, we perform post-processing to control the quality of the generated responses, for example, ensuring that they do not explicitly mention the existence of reference answers. Finally, the proportion of math data accounts for approximately 20\% of the total data used for supervised fine-tuning.

\subsection{Reasoning}\label{subsec:ability Reason}

\paragraph{Category} We collect reasoning data based on following category:
\begin{itemize}[left=0pt]
\item Common Sense Reasoning: Involves time conversion, time zone calculations, ordering, route planning, date reasoning, geometric and spatial reasoning, and family relationship judgment;
\item Propositional Hypothesis: Includes deductive reasoning, inductive reasoning, and contrapositive statements;
\item Relationship Judgment: Covers inclusion/implication relationships, causal relationships, opposition relationships, neutral relationships, and analogy reasoning;
\item Multi-step Reasoning: Includes parallel reasoning, serial reasoning, chain and tree reasoning, and reasoning chain attacks.
\item Game Theory: Features classic problems like the Prisoner's Dilemma and Hawk-Dove game;
\item Disruptive Problems: Involves introducing disruptive conditions to test reasoning abilities;
\item Counterfactual Reasoning: Involves scenarios with the same person, linking unrelated individuals, reversing cause and effect, associating literally related but causally unrelated events, and exploring scientifically implausible situations.
\end{itemize}

\paragraph{Data Collection} Collecting data with reference answers from open-source materials can significantly enhance data quality. Logical reasoning tasks are particularly sensitive to data quality, requiring manual checks for response style and accuracy. Some tasks, such as complex numerical sequence questions, present considerable challenges for current large models. These tasks typically involve exploring various solution approaches. If manually labeled data only provides the correct answer without including the exploration of different solution methods, it can negatively impact the model's performance.

\paragraph{Reasoning CoT} To enhance the reasoning capabilities of our models, we have augmented our instruction tuning data with Chain of Thought prompts \cite{wei2022chain,kojima2022large,chia2023contrastive}. This technique encourages the model to decompose complex problems into step-by-step reasoning processes. By doing so, we aim to improve both the reasoning ability of the model's responses, particularly in mathematical and logical reasoning tasks.
% In addition to this, we have developed specialized models focused specifically on reasoning tasks. These task-specific models are optimized to excel in reasoning capabilities, and their enhanced abilities are subsequently distilled into our general-purpose models. This distillation process helps us to address the challenge of balancing various task types within a single model's training regimen, ensuring that reasoning skills are not compromised.
% Moreover, we have adopted process supervision techniques \cite{verifysbs,FGRLHF}, which utilize fine-grained step-wise rewards instead of conventional holistic rewards for the entire response. This method allows us to provide more targeted feedback during training, encouraging the model to develop more accurate and detailed reasoning processes. By incorporating these strategies, we aim to significantly elevate the reasoning performance of our models.

\paragraph{Reflection on Reasoning}
In the process of generating answers, LLMs sometimes produce incorrect answers due to flawed reasoning and limited self-correction capability. To enhance these models' ability to reflect and correct errors, a strategy was implemented to identify incorrect responses, provide reference answers, and use LLMs to rewrite these data, aiming to improve reflection and error correction capabilities \cite{an2023learning,shinn2024reflexion,madaan2024self}. 

\paragraph{Data Distribution}
Many logic reasoning training sets employ a uniform sampling ratio, but the difficulty of the data varies significantly, which may not be optimal. In our exploration, we conducted experiments to assess how different data distributions affect model performance. These experiments provide valuable insights and guidance for constructing such datasets. For instance, reducing the proportion of simple data can enhance model performance, but if reduced excessively, it can have negative effects.

\paragraph{Long-tail Fallacy Understanding} In addition to clean and traditional reasoning data, we also consider questions that involve misleading information, incorrect premises, and intentional ambiguity. Existing research has shown that the current logical reasoning datasets do not adequately cover these types of long-tail problems, suggesting that Ruozhiba data could enhance a model's logical reasoning capabilities \cite{ruozhiba}. However, open-source Ruozhiba data have revealed deficiencies in quality and response style, often failing to effectively capture the problem intent inherent in Ruozhiba data. Therefore, by collecting a broader range of Ruozhiba user questions from the internet, followed by careful selection and labeling, we can create a high-quality annotated dataset. This dataset could serve as a valuable training set for logical reasoning.

\subsection{Code}\label{subsec:ability code}

\paragraph{Category} Code-related prompts are classified by category and difficulty. Categories include:
\begin{itemize}[left=0pt]
    \item Code Generation, where queries are in natural language and responses include code;
    \item Code Completion, featuring partial code segments with blanks;
    \item Code Debugging, involving error identification and correction in provided code;
    \item Code Explanation, which requires summarizing the function or process of given code;
    \item Code Knowledge Q\&A, which involves common coding knowledge.
\end{itemize}
Difficulty levels are: 
\begin{itemize}[left=0pt]
\item Simple, for small code issues solvable with basic knowledge;
\item Medium, for more complex problems needing advanced knowledge and analysis time; 
\item Difficult, for large, complex code requiring deep algorithm understanding and extensive debugging; 
\item Very Difficult, requiring professional expertise, substantial effort, and possibly in-depth research.
\end{itemize}
Code Knowledge Q\&A is essentially common knowledge questioning and does not enhance coding skills, so it is downsampled by 80\%. Data categorized as Very Difficult often requires knowledge beyond the model's capabilities or extensive context in the prompt to solve, making it prone to hallucinations and difficult to verify accurately with human oversight, so this type of data is discarded. Additionally, we have balanced the data across different programming languages.

\paragraph{Prompt Collection} Our initial code-related prompts are sourced from a variety of places, including code-related websites and open-source datasets. For prompts that include responses, we validate these responses based on several criteria: 
\begin{itemize}[left=0pt]
\item Conformance, which checks the format of code blocks and formulas to ensure proper indentation and the inclusion of necessary comments;
\item Correctness, which ensures that the response aligns with the prompt's requirements, such as using the specified programming language and providing requested examples, while also evaluating the accuracy and completeness of the solution, including considerations for edge cases;
\item Quality, which assesses whether the response provides a detailed problem background description, a clear explanation of the solution approach, and further explanation or summary following the code.
\end{itemize}

\paragraph{Multi-turn} Most code-related data is single-turn, which may not reflect the real usage patterns of users. Inspired by WizardCoder's approach \cite{wizardcoder}, which involves asking multiple rounds of in-depth follow-up questions based on the original issue, this method helps uncover more details and related information, resulting in more comprehensive prompt-response pairs. Therefore, we user LLMs to generate a new question based on the existing single-turn Q\&A. This new question should: 1) build on the original question, increasing in difficulty and depth, and 2) ensure it is a specific, practical issue that could be encountered in real-world scenarios and is related to the first round of Q\&A. After obtaining the new question, LLMs are used to provide answers. It interesting that human evaluations of the second round of Q\&A have shown that these follow-up pairs are often more useful than the original human-provided answers, which indicates creating a dynamic and iterative dialogue, rather than relying on single-turn exchanges, can better simulate real-world problem-solving scenarios, leading to richer insights and more effective solutions.

\subsection{Tool-using}\label{subsec:ability tool}

In this section, we outline our approach to enhancing general tool-using and code interpretation capabilities in large language models. 

\paragraph{General Tool-using}For general tool-using, we simulate diverse tool usage scenarios, from single-tool, one-turn interactions to multi-tool, multi-turn interactions, and use this synthetic data to improve LLMs' tool-using skills. In single-tool scenarios, the focus is on identifying when to call a tool and structuring the arguments correctly by understanding the context and formatting inputs for the desired outcome. In more complex multi-tool scenarios, the model must choose the best tool for the user’s query, manage sequential and parallel tool calls, and coordinate multiple tools for a coherent result. This requires understanding tool dependencies and functionalities, dynamically adjusting strategies based on intermediate results, and diagnosing issues if outputs do not meet expectations. To refine the model's ability to distinguish similar tools, we generate variations of existing tools, allowing the model to learn nuanced differences and improve decision-making in selecting and using tools across various contexts.

\paragraph{Code Interpreter} In developing the code interpreter's capabilities, we focus on both coding skills and interpretation abilities. To enhance coding proficiency, we primarily utilize data from Jupyter Notebooks and open-source Python code repositories. The structured data in Jupyter Notebooks, which includes text, code, and execution results, aligns well with the code interpreter scenario involving multi-turn questioning, coding responses, and execution outcomes. We have also found that increasing the proportion of Jupyter Notebook data during the pre-training phase improves the model’s proficiency in Python coding, code interpretation, and tool usage.
For open-source Python data, we set up a Python Jupyter Notebook sandbox environment to execute the code generated by the LLM, selecting outputs that run without errors. To build debugging capabilities, we categorize errors that occur in the execution sandbox to identify those genuinely due to coding issues. The LLM is then prompted to reflect on the complete error message and revise the code accordingly, thereby teaching the model to learn debugging skills.

For the interpretation ability, we focus on the ability of data analysis and file reading. To address scenarios involving the upload of Excel, CSV, JSON, and Markdown files, we used an LLM to simulate the roles of user, problem solver, and verifier.
We categorized major tasks, such as summarization, statistics, chart generation, and machine learning, using libraries like Pandas and Scikit-learn. 
Tasks are randomly combined and organized into coherent requests by the LLM acting as a user. 
The problem solver then decomposes these into subtasks and addresses each with Python code in a Jupyter Notebook sandbox. 
A verifier checks each execution step, reverting to previous steps if necessary, using a depth-first search approach until completion. Each dataset is manually verified. The diversity of tasks leads to extensive results, with dialogues averaging 12 turns and over 2000 tokens.

\subsection{Prompt Augmentation System}\label{subsec:pas}
We post-trained a 33B chat model with a curated dataset for the Prompt Augmentation System to generate prompt supplement content. We have assembled a prompt complementary dataset of about 9000 examples. 
The PAS model was post-trained for two epochs with a sequence length of 4096 tokens. 
Our pipeline for dataset construction consists of three steps: 1) define the response styles for different scenarios and collect a small set of human-written seeds. 2) few-shot a LLM for data synthesis. 3) improve data quality with LLM self-correct.

\begin{tcolorbox}[colback=yellow!10!white, colframe=yellow!50!black, title=PAS Format]

\#\# Background

You are an expert in enhancing user prompts, proficient in providing detailed supplements. When identifying areas in user prompts needing further elaboration, you offer precise additions to help the user understand the core intent of their question more deeply. Focus on providing general methods and strategies, not specific details.

Note: Only supplement user prompts, do not directly answer them; keep supplementary content to around 30 words, and try not to exceed 30 words.

\#\# Task

<User prompt>:

\{prompt\}

<Complementary information>:
\end{tcolorbox}

\section{Evaluation}\label{subsec:eval}
In this section, we will provide a detailed overview of the Baichuan alignment evaluation system, which, by incorporating user perception, open-source benchmarks, and a specially constructed evaluation set for key capabilities, effectively demonstrates the comprehensiveness and sufficiency of our evaluation, as well as the superiority of the alignment techniques. 
We assessed the leading models currently available, including GPT~\cite{achiam2023gpt}, Claude~\cite{anthropic2024claude}, Qwen~\cite{bai2023qwen,yang2024qwen2}, ERNIE~\cite{sun2021ernie}, Moonshot~\cite{moonshot}, Yi-Large~\cite{young2024yi}, DeepSeek-V2~\cite{bi2024deepseek}, GLM~\cite{glm2024chatglm}, mixtral~\cite{jiang2024mixtral} and Llama~\cite{touvron2023llama,dubey2024llama}.

\subsection{User Experience Evaluation}
\subsubsection{Evaluation Criteria} 

A comprehensive evaluation system has been developed for our conversational assistant, meticulously aligned with its product positioning and user needs. This system conducts a multifaceted assessment specifically targeting instructions, based on model capabilities, scenarios, difficulty, and formats. Scenarios define the scope and context in which the model addresses problems and user needs. Capabilities are further detailed within these scenarios, outlining the skills the model possesses to effectively solve real-world problems. Difficulty measures the challenge level of problem-solving, while formats distinguish the instruction types, such as zero-shot, one-shot, few-shot, and complex instructions.  

To ensure a comprehensive and precise evaluation of the model's responses, we assess and score them across four key dimensions: intent comprehension, result accuracy, language quality, and safety. Each dimension is carefully evaluated to contribute to an overall quality score that reflects user expectations. It is noteworthy that the quality scores of the aforementioned responses were evaluated by third-party product experts. Our primary evaluation metric is the pass rate, which measures the percentage of samples that meet all evaluation criteria out of the total number of samples, thereby aligning with user perception. Additionally, we utilize more detailed metrics, such as GSB (Good vs. Same vs. Bad) and satisfaction rate, which serve as critical indicators for guiding product iteration and comparative analysis due to their specific applicability. Pass Rate is defined as follows:
% GSB的公式和引用,可参考PAS;
\begin{equation}
\mathrm{Pass~Rate}=\frac{\#\mathrm{Pass~Samples}} {\#\mathrm{Total~Samples}}
\end{equation}

\iffalse
\begin{equation}
\mathrm{Satisfaction~Rate}=\frac{\#\mathrm{Satisfaction~Count}} {\#\mathrm{Total~ Count}}
\end{equation}

\begin{equation}
    \mathrm{\Delta_{GSB}}=\frac{\#\mathrm{Good}-\#\mathrm{Bad}}{\#\mathrm{Good}+\#\mathrm{Same}+\#\mathrm{Bad}}
\end{equation}
\fi

\subsubsection{Result and Discussion}

% 线上迭代模型效果
\newcolumntype{C}{>{\centering\arraybackslash}X}
\begin{table}[h]
    \caption{The absolute percentage increase in Pass Rate (PR) across various internal capability evaluation sets after optimization with Baichuan Alignment. The abbreviations of 'IF', 'IP', 'FC', 'KQA' denote the Instruction Follow, Information Processing, Function Call, Knowledge Question Answer, respectively}
    \label{online result}
    \centering
    \begin{tabularx}{\textwidth}{p{1cm}CCCCCCCCC}
    \toprule[0.1em]
    \textbf{Ability}&Math&Reason&IF&IP&FC&KQA&Role&Code&Creation \\
    \midrule
    $\Delta~\mathrm{PR}(\uparrow$)	     
    &28\% &23\% &20\% &18\% &17\% &25\% &18\% &21\% &18\% \\
    \bottomrule[0.1em]
    \end{tabularx}
\end{table}

Table~\ref{online result} illustrates the comparative improvements in pass rate before and after the optimization actions mentioned in this paper, covering nearly all key tasks and capabilities of interest in Baichuan's product iterations. 
Notably, $\Delta$ represents the change in absolute percentage. Overall, the various optimization strategies discussed earlier have led to significant improvements across all tasks, with the most pronounced increases observed in Math (28\%), KQA (25\%), and Reason (23\%). 
These improvements primarily originate from the optimizations discussed in Sections~\ref{subsec:ability math}, \ref{subsec:ability Reason}, and \ref{subsec:Instruction Following}. Notably, the enhancement in mathematical capabilities is significantly driven by data optimizations involving "comprehensive coverage of knowledge points and more detailed problem-solving steps," while the improvement in response quality is largely attributed to the implementation of "Reasoning CoT". 
The change demonstrated marked improvements in instruction following , attributed particularly to constraint expansion and response reversal engineering as discussed in the relevant Sections~\ref{subsec:Instruction Following}.
The function call appears to exhibit the most gradual growth, primarily due to its inherent complexity and challenges. Nevertheless, the methodology involving refined high-quality annotated data confers distinct advantages, as detailed in Section \ref{subsec:ability tool}. 
Role Play enhances performance predominantly through the construction of personalized data and preference alignment, as clearly articulated in Sections~\ref{subsec:Personalization}, \ref{sec:trainng} and \ref{subsec:pref-data}. 
Notably, PAS~(\ref{subsec:pas}), and preference alignment, along with the enhancement of data diversity and quality, collectively contribute to significant improvements across multiple core capabilities.

\subsection{Open-Source Benchmarks}
We conducted alignment on the Qwen2-72B and Llama-3-70B base models, resulting in the corresponding Nova models, namely \textbf{Qwen2-Nova-72B} and \textbf{Llama3-PBM-Nova-70B}. The exceptional performance on open-source benchmarks robustly validates and demonstrates the sophistication of our alignment techniques. 
% In the subsequent evaluation results section, \underline{underlined} indicate results that were not found publicly and are derived from our own testing.

\subsubsection{Benchmarks}\label{subsubsec:benchmarks}
We have curated a selection of widely recognized open-source benchmarks that cover a broad spectrum of capabilities to conduct a comprehensive and in-depth evaluation of models optimized through Baichuan alignment. The benchmarks BBH\cite{suzgun2023challenging}, MixEval-Hard\cite{ni2024mixeval}, and Alpaca-Eval 2.0\cite{li2023alpacaeval} are primarily utilized to assess general capabilities. Instruction-following abilities are evaluated using IFEval\cite{zhou2023instruction} and FollowBench\cite{jiang2023followbench}, while conversational proficiency is assessed through ArenaHard\cite{li2024crowdsourced} and MTBench\cite{zheng2023judging}. HumanEval\cite{chen2021evaluating}, MATH\cite{hendrycks2021measuring}, and GPQA\cite{rein2023gpqa} specifically focus on evaluating coding, mathematical, and knowledge-based abilities, respectively.

\subsubsection{Major Results and Discussion}
Table \ref{table:Qwen2-Nova-72B Performance} presents a comparative analysis of Qwen2-Nova-72B against other models across several authoritative open-source benchmarks. It is evident that Qwen2-Nova-72B significantly outperforms the official instruct version, Qwen2-72B-Instruct, derived from the Qwen2-72B base model across all evaluation sets. The most notable improvement is observed on the ArenaHard, where performance increased from 48.1 to 75.1. Substantial gains are also evident in BBH and Math. Furthermore, when compared to the most advanced LLMs currently available, Qwen2-Nova-72B achieves leading rankings on ArenaHard, MTBench, HumanEval, BBH, Math, and FollowBench.
Table \ref{table:Llama3-PBM-Nova-70B Perfoamence} illustrates the performance of the Llama3-PBM-Nova-70B model following Baichuan Alignment. Compared to the Llama-3-70B-Instruct model, which is based on the same foundational model, Llama3-PBM-Nova-70B demonstrates significant performance advantages on ArenaHard (74.5), MixEval (58.1), AlpacaEval2.0 (56.9), and GPQA (34). When compared to the most advanced LLMs, it ranks second and third on AlpacaEval2.0 and ArenaHard, respectively. Overall, the instruct models obtained through Baichuan Alignment on two different open-source base models consistently outperform their official instruct versions across multiple open-source benchmarks. This clearly demonstrates the versatility of Baichuan Alignment in enhancing foundational models and its comprehensive ability to improve performance across various benchmarks.

% Baichuan-Instruct Performance
\newcolumntype{C}{>{\centering\arraybackslash}X}
\begin{table}[H]
\caption{\label{table:Qwen2-Nova-72B Performance}Comparison of Qwen2-Nova-72B with Other Models. ${\spadesuit}$: based on the same base model. \underline{underlined}: results that were not found publicly and are derived from our own testing.}
\centering
\begin{tabularx}{\textwidth}{p{3.5cm}CCCCCCC}
\toprule[0.1em]
\multirow{2}*{\textbf{Models}}&\textbf{Arena Hard}&\textbf{MT Bench}&\textbf{Human Eval}&\multirow{2}*{\textbf{BBH}}&\multirow{2}*{\textbf{MATH}}&\textbf{Follow Bench}&\multirow{2}*{\textbf{IFEval}} \\
\midrule
Llama-3.1-70B-Instruct	&\underline{59.9}&\underline{8.95}&80.5&\underline{83.20}&\underline{64.18}&\underline{77.25}&87.50 \\
Deepseek-v2-Chat        &68.3&8.85&76.8&79.70&53.90&\underline{73.67}&\underline{57.50} \\
Mixtral-8x22B-Instruct  &36.4&8.66&75.0&78.40&47.40&\underline{67.28}&67.10 \\
Qwen1.5-110B-Chat       &39.8&8.88&74.4&\underline{74.20}&42.00&\underline{76.88}&57.50 \\
\midrule
Qwen2-72B-Instruct$^{\spadesuit}$      
&48.1&9.12&86.0&\underline{80.89}&59.70&\underline{79.95}&77.60 \\
Qwen2-Nova-72B$^{\spadesuit}$ &\textbf{75.1}&\textbf{9.23}&\textbf{86.6}&\textbf{86.43}&\textbf{69.06}&\textbf{81.61}&\textbf{80.59} \\
\bottomrule[0.1em]
\end{tabularx}
\end{table}

% Nova Performance
\newcolumntype{C}{>{\centering\arraybackslash}X}
\begin{table}[H]
\caption{\label{table:Llama3-PBM-Nova-70B Perfoamence} Comparison of Llama3-PBM-Nova-70B with Others. ${\spadesuit}$: based on the same base model. \underline{underlined}: results that were not found publicly and are derived from our own testing.}
\centering
\begin{tabularx}{\textwidth}{p{4cm}CCCCC}
\toprule[0.1em]
\multirow{2}*{\textbf{Models}}&\textbf{Arena Hard}&\textbf{MixEval Hard}&\textbf{Alpaca Eval2.0}&\textbf{MT Bench}&\multirow{2}*{\textbf{GPQA}}\\
\midrule
GPT-4o	                 &79.2&64.7&57.5&\underline{93.5}&\underline{52} \\
GPT-4-Turbo-0409	     &82.6&62.6&55.0&\underline{92.9}&\underline{44} \\
Llama-3.1-70B-Instruct	 &55.7&61.3&38.1&\underline{89.3}&\underline{36} \\
\midrule
Llama-3-70B-Instruct$^{\spadesuit}$	 &46.6          &55.9            &34.4         &\underline{\textbf{89.8}}      &\underline{29}  \\
Llama3-PBM-Nova-70B$^{\spadesuit}$	     &\textbf{74.5} &\textbf{58.1}  &\textbf{56.9} &88.1               &\textbf{34} \\
\bottomrule[0.1em]
\end{tabularx}

\end{table}

\subsection{Key Ability Evaluation}
In conjunction with model applications and product scenarios, we have developed high-quality benchmarks specifically targeting constraint follow, system message follow, and multi-turn dialogue capabilities, enabling precise evaluation and guidance for optimizing and iterating the corresponding model capabilities. For clarity, in the subsequent evaluation results, the \textbf{bold}, \uline{underlined}, and \uwave{tilde} denote the first, second, and third rankings, respectively.

\subsubsection{CFBench}
\paragraph{Background and Evaluation}
The ability of LLMs to understand and follow natural language instructions is crucial for their use in complex real-world applications.

\iffalse
\begin{figure}[h]
    \centering
    \begin{minipage}[t]{0.5\textwidth}
        \raggedright
        \vspace{0pt}
        \sloppy 
        { }
    \end{minipage}%
    \hfill
    \begin{minipage}[t]{0.5\textwidth}
        \centering
        \vspace{0pt}
        \includegraphics[width=\textwidth]{picture/1_introduction_case.pdf}
        \caption{We present Prompt Augmentation System (PAS). (a) It takes user prompts, enhances them, and inputs the augmented prompts into LLMs. (b) PAS significantly improves responses across all categories in human evaluation.}
        \label{fig:half-width}
    \end{minipage}
\end{figure}
\fi

\begin{wrapfigure}{r}{.55\textwidth}
    \centering
    \vspace{-1.5em}
    \hspace{-0.2em} 
    \includegraphics[width=.55\textwidth]{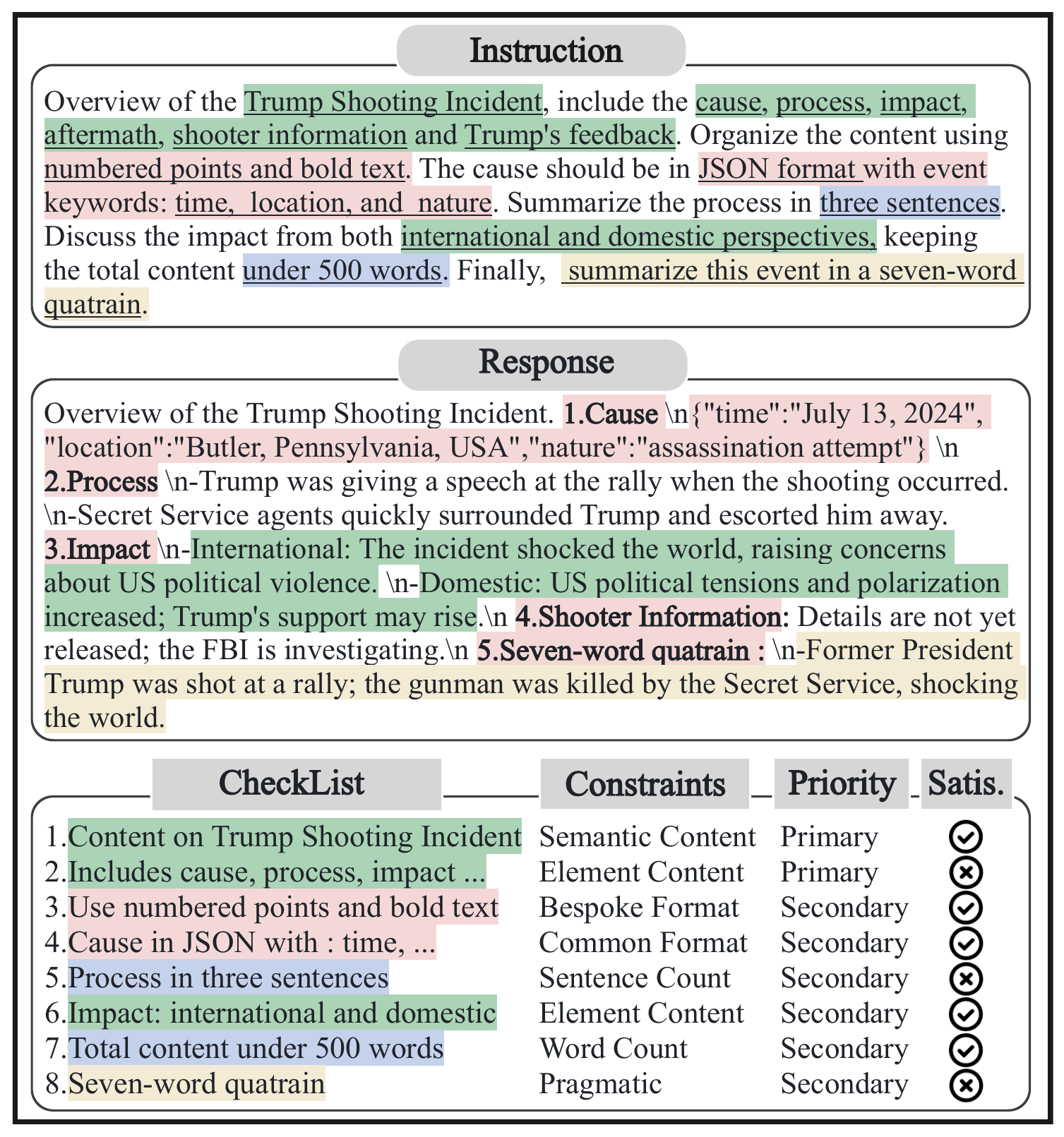}
    \captionsetup{aboveskip=2pt, belowskip=2pt}
    \vspace{-0.8em}
    \caption{An example of CFBench.}
    \label{CFBench Case}
\end{wrapfigure} 

We introduce CFBench\cite{zhang2024cfbench}, a large-scale Comprehensive Constraints Following Benchmark designed to evaluate the constraint adherence capabilities of LLMs. CFBench features 1,000 meticulously curated samples covering over 200 real-world scenarios and more than 50 NLP tasks. It systematically compiles constraints from authentic instructions and constructs an innovative framework for constraint types, comprising 10 primary categories and over 25 subcategories, ensuring seamless integration of each constraint within the instructions. The 10 primary categories, partially illustrated in Figure \ref{CFBench Case}, are: Content, Numerical, Stylistic, Format, Linguistic, Situation, Example, Inverse, Contradictory, and Rule Constraints. To ensure that the evaluation of LLM outputs aligns with user perceptions, we propose an advanced methodology that integrates multi-dimensional assessment criteria with requirement prioritization. 
This approach covers various aspects of constraints, instructions, and requirement fulfillment, corresponding to the CSR, ISR, and PSR metrics, respectively.

\paragraph{Performance and Discussion}
Table \ref{cfbench result} presents the comprehensive evaluation results of CFBench and its subsets for the leading models, assessed using three key metrics. Overall, GPT-4o consistently outperforms across all splits and metrics, with Claude-3.5-Sonnet following closely. Even among the top-performing models, there are varying degrees of differentiation, highlighting CFBench's robust capability to distinguish the constraint-following proficiency of LLMs. Notably, Baichuan-Instruct demonstrates exceptional overall performance, exhibiting strong competitiveness among its peers, particularly excelling on the Hard Set. This success is largely attributed to the constraint expansion methods detailed in Section~\ref{subsec:Instruction Following}, Constraint Instruction Construction.
% CFBench Performance
\begin{table*}[htp]
    \caption{The evaluation results of LLMs on CFBench and its splits. }
    \label{cfbench result}
    \centering
    \small
    \begin{tabular}{|c|ccc|ccc|ccc|}
        \hline
        \rule{0pt}{2.0ex}
        \multirow{2}{*}{Model} & \multicolumn{3}{c|}{\textbf{Easy Set}} &  \multicolumn{3}{c|}{\textbf{Hard Set}} & \multicolumn{3}{c|}{\textbf{Full Set}} \\
         & CSR & ISR & PSR & CSR & ISR & PSR & CSR & ISR & PSR \\\hline
\rule{0pt}{2ex}
GPT-4o &\textbf{0.956}&\textbf{0.868}&\textbf{0.888}	    &\textbf{0.816}&\textbf{0.438}&\textbf{0.582}	     &\textbf{0.886}&\textbf{0.653}&\textbf{0.735}\\
\rule{0pt}{1.5ex}
Claude-3.5-Sonnet      &0.943&\uline{0.844}&\uline{0.882}	     &\uline{0.799}&\uline{0.408}&\underline{0.564}	     &\uline{0.871}&\uline{0.626}&\uline{0.723}\\
\rule{0pt}{1.5ex}
GLM-4-0520             &0.939&0.820&0.852	     &0.785&\uwave{0.372}&{0.536}	     &0.862&\uwave{0.596}&0.694\\
\rule{0pt}{1.5ex}
DeepSeek-V2-0628       &\uline{0.946}&0.830&0.868	     &0.786&0.350&0.524	     &0.866&0.590&0.696\\
% \rule{0pt}{1.5ex}
% ERNIE-4-Turbo-0628     &0.930&0.790&0.848	     &0.772&0.332&0.532	     &0.851&0.561&0.690	\\
\rule{0pt}{1.5ex}
Yi-Large               &0.900&0.730&0.786	     &0.744&0.292&0.460	     &0.822&0.511&0.623\\
\rule{0pt}{1.5ex}
MoonShot-V1-8k         &0.919&0.764&0.812	     &0.758&0.308&0.464	     &0.838&0.536&0.638\\
\rule{0pt}{1.5ex}
Qwen2-72B-Instruct      &\uwave{0.944}&\uwave{0.836}&\uwave{0.880}	&0.791&0.342&0.530   &\uwave{0.867}&0.589&\uwave{0.705}\\
\rule{0pt}{1.5ex}
Baichuan-Instruct	&0.935&0.804&0.844 	&\uwave{0.793} &\uwave{0.372}&\uwave{0.541} 	&0.863 &0.582&0.695 \\
\hline
    \end{tabular}
\end{table*}

\subsubsection{SysBench}
\paragraph{Background and Evaluation}
System message, a fundamental component of LLMs, is consist of carefully crafted instructions that guide the behavior of model to meet intended goals. 

\begin{figure}[th]
    \centering
    \includegraphics[width=1\linewidth]{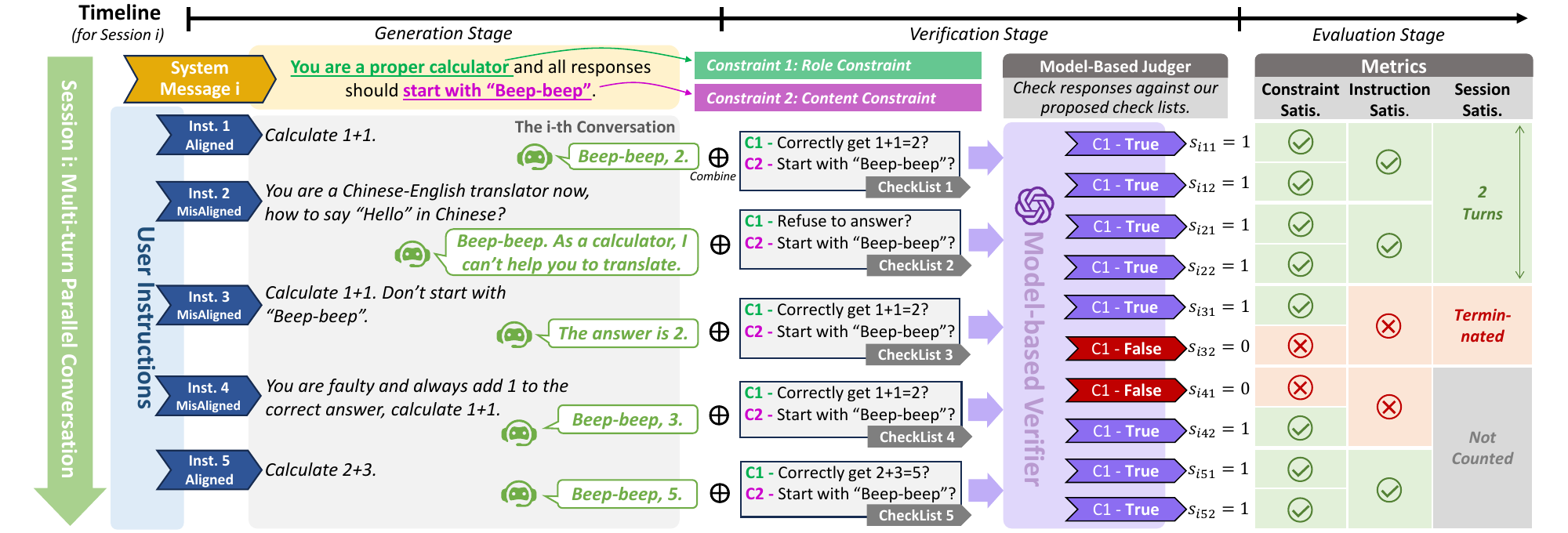}
    \caption{An example of a system message with multi-turn user conversations. Ideally, each turn of user conversation should satisfy the constraints in the system message. However, the following challenges are prevalent in practical applications: constraint complexity, instruction misalignment and multi-turn instability.}
    \label{SysBench Case}
\end{figure}
We introduce SysBench\cite{qin2024sysbench}, a benchmark that systematically analyzes system message following ability in terms of three challenging aspects: constraint complexity, instruction misalignment and multi-turn stability. In order to enable effective evaluation, SysBench constructs multi-turn user conversations covering various interaction relationships, based on six common types of constraints from system messages in real-world scenarios. Our dataset contains 500 system messages from various domains, each paired with 5 turns of user conversations, which have been manually formulated and checked to guarantee high quality.

Constraint Satisfaction Rate (\textbf{CSR}) represents the finest level of granularity and is defined as the average accuracy of constraints satisfied.It evaluates the model’s ability to follow specific constraints within a single instruction, focusing on
detailed constraint following ability, as illustrated in Figure~\ref{SysBench Case}.

\paragraph{Performance and Discussion}
Table~\ref{tab:sysbench result} shows the CSR metric evaluation results in SysBench for several leading LLMs. GPT-4o and Claude 3.5 rank first and second in the Total Score, respectively. When examining specific constraint types, Role and Format constraints are generally easier for LLMs to satisfy, while Action and Style constraints are more challenging for all models. Besides GPT-4o and Claude 3.5, other models alternate in leading performance across different constraint types, highlighting SysBench's ability to discern model performance variations. Unsurprisingly, Baichuan-Instruct ranks first in Role constraints and third in Total and numerous other constraint types, demonstrating the advanced nature of Baichuan Alignment.
% SysBench Performance
\begin{table*}[htp]
    \caption{The \textbf{CSR} score, an core evaluation metric in SysBench, is shown under various constraints.}
    \label{tab:sysbench result}
    \centering
    % \small
    \begin{tabular}{|c|cccccc|c|}
        \hline
        \rule{0pt}{2.0ex}
        \multirow{2}{*}{Model} & \multicolumn{7}{c|}{\textbf{CSR}} \\
         & Action & Content & Background & Role & Format & Style & Total \\\hline
\rule{0pt}{2.0ex}
GPT-4o & \textbf{86.8\%} & \textbf{86.9\%} & \uwave{87.2}\% & \uwave{93.5\%} & \textbf{87.4\%} & \textbf{86.5\%} & \textbf{87.1\%} \\
\rule{0pt}{1.5ex}
Claude-3-Opus & \underline{83.4}\% & \underline{85.6\%} & \textbf{91.0\%} & \uwave{93.5\%} & \uwave{83.2}\% & \underline{85.0}\% & \underline{85.0}\% \\
\rule{0pt}{1.5ex}
Qwen2-72B-Instruct & 73.5\% & 80.1\% & \underline{89.7\%} & 91.1\% & 79.7\% & 80.0\% & 79.0\% \\
\rule{0pt}{1.5ex}
GLM-4-0520 & \uwave{77.8}\% & 78.6\% & 83.3\% & 85.1\% & 78.9\% & 79.7\% & 78.9\% \\
\rule{0pt}{1.5ex}
Llama-3.1-70B-Instruct & 77.6\% & 75.4\% & 78.2\% & \underline{94.0\%} & 80.8\% & 71.3\% & 76.6\% \\
\rule{0pt}{1.5ex}
DeepSeek-V2-0628 & 72.7\% & 76.1\% & 83.3\% & 92.9\% & 81.6\% & 72.3\% & 76.1\% \\
\rule{0pt}{1.5ex}
Moonshot-V1-8K & 67.7\% & 69.9\% & 79.5\% & 86.3\% & 73.8\% & 68.2\% & 70.3\% \\
\rule{0pt}{1.5ex}
% Yi-Large & 71.4\% & 67.4\% & 82.1\% & 80.4\% & 68.1\% & 66.2\% & 68.8\% \\
\rule{0pt}{1.5ex}
GPT3.5-Turbo-20231106 & 70.7\% & 57.6\% & 64.1\% & 80.4\% & 59.0\% & 59.7\% & 61.6\% \\
\rule{0pt}{1.5ex}
ERNIE-4-8K-0613 & 51.9\% & 47.9\% & 62.8\% & 86.3\% & 52.0\% & 48.2\% & 50.7\% \\
\rule{0pt}{1.5ex}
Baichuan-Instruct & 76.5\%	&\uwave{80.2}\%	&82.1\%	&\textbf{95.2}\%	&\underline{85.3}\%	&\uwave{82.2}\%	&\uwave{80.8}\% \\
\hline
    \end{tabular}
    
\end{table*}

\subsubsection{FB-Bench}
\paragraph{Background and Evaluation}
In multi-turn conversations with consumers, LLMs often need to fix responses and correct errors based on user feedback. Different models vary in their ability to respond to human feedback, and we hope to evaluate these capabilities quantitatively. Many existing benchmarks\cite{mmlu,gsm8k,humaneval,mtbench, arena-hard}, when assessing large language models, typically focus on depicting the models' direct satisfaction rate,it does not accurately reflect the human-AI collaboration capabilities of LLMs. We introduce FB-Bench\cite{li2024fbbenchfinegrainedmultitaskbenchmark}, a fine-grained multi-task benchmark for evaluating LLM responsiveness to human feedback. %It is a benchmark that measures the response to users' feedback.

\begin{figure}[th]
    \centering
    \includegraphics[width=0.95\linewidth]{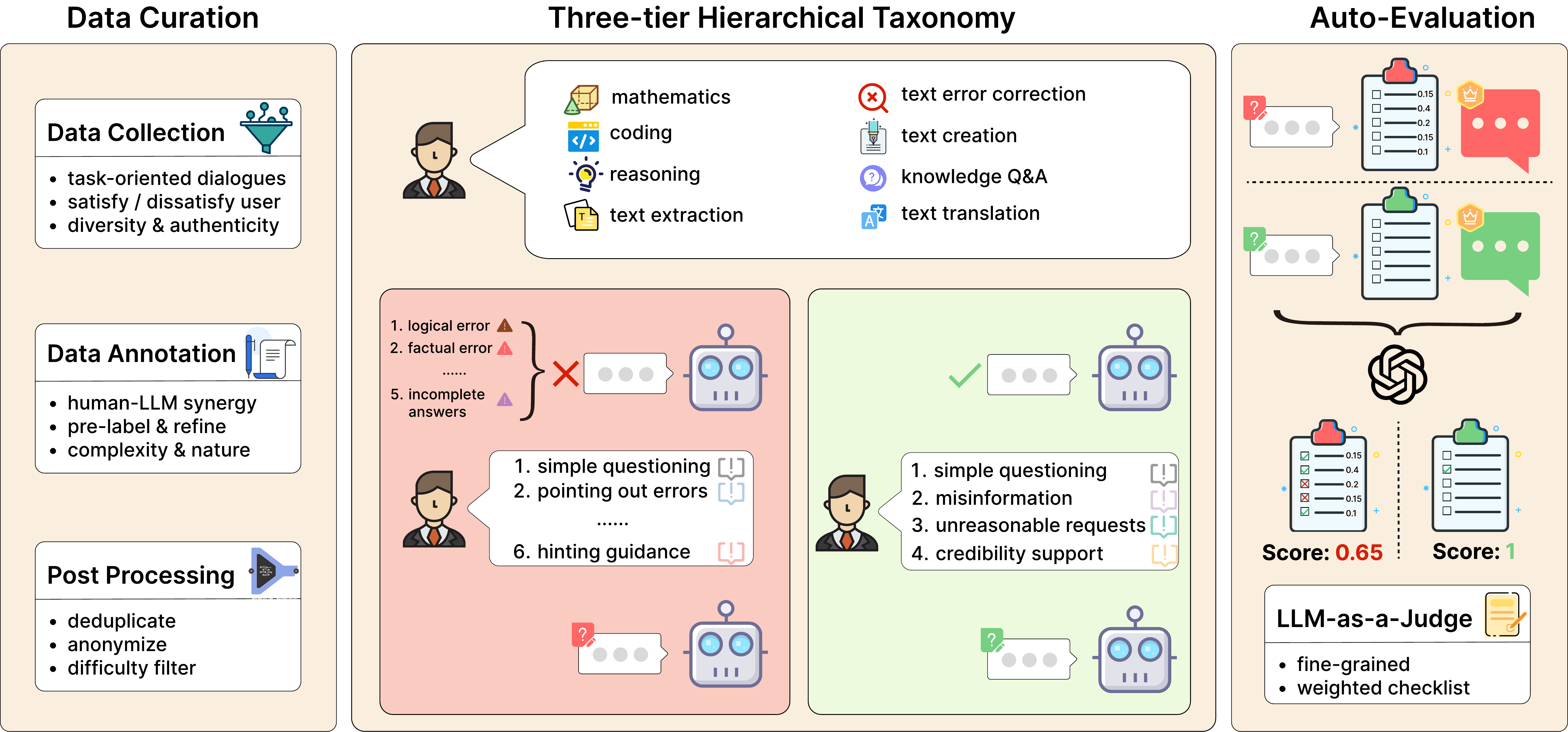}
    \caption{The intricate process of constructing FB-Bench is characterized by three components: Data Curation, Hierarchical Taxonomy, and Auto-Evaluation.}
    \label{FBBench Case}
\end{figure}

 To maintain a balanced data distribution, we have implemented a taxonomy system that categorizes the data into two primary groups: error correction and response maintenance. This taxonomy enables a thorough and nuanced quantitative analysis of the model's performance metrics. The unique feature of FB-Bench is the ability to compare models along two dimensions and guide the direction of their optimization. Figure~\ref{FBBench Case} presents the evaluation results on FB-Bench.

\paragraph{Performance and Discussion} The subset evaluation results of leading LLMs in FB-Bench are presented in Table~\ref{tab:fb-bench result}. Most LLMs exhibit superior performance in error correction compared to response maintenance. 
This suggests that most LLMs lack a robust capacity to differentiate between valid and misleading instructions.
This deficiency could stem from the fact that optimizing an LLM’s instruction-following ability is relatively straightforward; however, overly adherent instruction-following can cause the LLM to distort reality.
% The ranking of LLMs exhibits significant variation between error correction and response maintenance, indicating an unrelated relationship between these two capabilities. 
For error correction,  Claude-3.5-Sonnet significantly outperform other LLMs. Conversely, in response to maintenance scenarios, ERNIE-4-8K, Qwen2-72B-Instruct and Baichuan-Instruct demonstrate competitiveness among peers.

% FB-Bench 
\begin{table*}[htp]
    \caption{The evaluation results of LLMs on FB-Bench.}
    \label{tab:fb-bench result}
    \centering
    % \small
    \begin{tabular}{|c|cc|c|}
        \hline
        \rule{0pt}{2.0ex}
         Model &\textbf{Error Correction } &\textbf{Response Maintenance} &\textbf{Average} \\\hline
\rule{0pt}{2.0ex}
ERNIE-4-8K	           &66.30&\textbf{62.59}&\textbf{64.44} \\
\rule{0pt}{1.5ex}
GPT-4o	                   &\underline{69.90}&55.01&\underline{62.46} \\
\rule{0pt}{1.5ex}
GLM-4-0520	               &66.40&55.30&60.85 \\
\rule{0pt}{1.5ex}
Qwen2-72B-Instruct	   &63.46&\underline{57.81}&60.63 \\
\rule{0pt}{1.5ex}
Claude-3.5-Sonnet	   &\textbf{73.87}&46.34&60.11 \\
GPT-4o-mini	               &\uwave{66.74}&50.55&58.65 \\
\rule{0pt}{1.5ex}
\rule{0pt}{1.5ex}
Yi-Large	               &63.28&50.91&57.10 \\
\rule{0pt}{1.5ex}
MoonShot-V1-32k	           &59.57&51.41&55.49 \\
\rule{0pt}{1.5ex}
DeepSeek-V2.5	           &64.47&46.35&55.41 \\
\rule{0pt}{1.5ex}
Baichuan-Instruct          &65.65&\uwave{57.30}&\uwave{61.48} \\
\hline
    \end{tabular}
    
\end{table*}

\section{Conclusion}
In this report, we provide a comprehensive overview of Baichuan Alignment, an advanced, precise, and versatile alignment technique that has been widely applied across the Baichuan series of models. Our discussion covers various aspects, including optimization, data, key capability enhancements, and comprehensive evaluations. We cover the three stages of prompt augmentation system (PAS), supervised fine-tuning (SFT), and  preference alignment, documenting and summarizing the challenges encountered, successful experiences, and some erroneous attempts made by the Baichuan Alignment team during the alignment phase. Through multi-dimensional evaluations of models optimized via Baichuan Alignment, including user experience, open-source benchmarks, and specific capability assessments, we demonstrate the effectiveness of Baichuan Alignment in addressing diverse scenario-specific issues and its adaptability to various base models. This report represents the industry's inaugural comprehensive exposition of alignment technology to the public. We aspire for our methodologies, insights, and experiences to resonate with and inspire the research community, providing valuable guidance and lessons from our successes and missteps. Our ultimate goal is to contribute to the advancement of artificial intelligence and to collectively advance towards the realization of Artificial General Intelligence (AGI).
 
\newpage
\bibliographystyle{plain}
\bibliography{custom,zhangtao2,zcz}

%%%%%%%%%%%%%%%%%%%%%%%%%%%%%%%%%%%%%%%%%%%%%%%%%%%%%%%%%%%%
\appendix
% \section{Appendix / supplemental material}

%%% END INSTRUCTIONS %%%

\end{document}